\documentclass{ecai}
\usepackage{times}
\usepackage{graphicx}
\usepackage{latexsym}
\usepackage{multirow}

\ecaisubmission   % inserts page numbers. Use only for submission of paper.
\usepackage{xcolor}                     % colors
\usepackage[utf8]{inputenc}             % Umlaute 
\usepackage[T1]{fontenc}                % Umlaute 
\usepackage{amsmath,amsfonts,amsthm}
\usepackage[ruled,vlined]{algorithm2e}
\usepackage{todonotes}
\usepackage[normalem]{ulem}
\usepackage{url}

\usepackage[export]{adjustbox}
\usepackage{hyperref}
\hypersetup{
      colorlinks = true,
      linkcolor = green!40!black,
      anchorcolor = green!40!black,
      citecolor = green!40!black,
      filecolor = green!40!black,
      urlcolor = green!40!black
      }

\usepackage[nameinlink,noabbrev]{cleveref}

\newcommand{\commentPC}[1]{{#1}}
\newcommand{\outPC}[1]{}
\newcommand{\commentHG}[1]{}
\newcommand{\outHG}[1]{}
\newcommand{\commentMR}[1]{{#1}}
\newcommand{\outMR}[1]{}

\newcommand{\outLRK}[1]{}

\DeclareMathOperator*{\argmax}{arg\, max}

\newcommand{\IoU}{\mathit{IoU}}
\newcommand{\mIoU}{\mathit{mIoU}}

\newcommand{\cA}{\mathit{cost}_A}
\newcommand{\cB}{\mathit{cost}_B}
\newcommand{\cP}{\mathit{cost}_P}

\begin{document}

% ----------------------------------------------------------------------------------------------------------------
% ------------------------------------             TITLE & ABSTRACT         --------------------------------------
\title{MetaBox+: A new Region Based Active Learning Method for Semantic Segmentation using Priority Maps}
% \title{Active Learning for semantic segmentation: region based, cost effective, meta regression.}
\author{Pascal Colling$^{1,2}$, Lutz Roese-Koerner$^{1}$, Hanno Gottschalk$^{2}$, Matthias Rottmann$^{2}$ \\
\\
$^{1}$Aptiv, Wuppertal, Germany\\
$^{2}$University of Wuppertal, School of Mathematics and Natural Sciences, IMACM  \& IZMD \\
\tt\small{\{\href{mailto:pascal.colling@aptiv.com}{pascal.colling},\href{mailto:lutz.roese-koerner@aptiv.com}{lutz.roese-koerner}\}@aptiv.com.de} \\
\tt\small{\{\href{mailto:hanno.gottsch@uni-wuppertal.de}{hanno.gottschalk},\href{mailto:rottmann@uni-wuppertal.de}{rottmann}\}@uni-wuppertal.de}
}

\maketitle

\begin{abstract}
We present a novel region based active learning method for semantic image segmentation, called \emph{MetaBox+}. For acquisition, we train a meta regression model to estimate the segment-wise Intersection over Union ($\IoU$) of each predicted segment of unlabeled images. This can be understood as an estimation of segment-wise prediction quality. Queried regions are supposed to minimize to competing targets, i.e., low predicted $\IoU$ values / segmentation quality and low estimated \commentMR{annotation} costs. For estimating the latter we propose a simple but practical method for \commentMR{annotation} cost estimation.
We compare our method to entropy based methods, where we consider the entropy as uncertainty of the prediction. The comparison and analysis of the results provide insights into annotation costs as well as robustness and variance of the methods.
Numerical experiments conducted with two different networks on the Cityscapes dataset clearly demonstrate a reduction of annotation effort compared to random acquisition. \commentMR{Noteworthily}, we achieve $95\%$ of the mean Intersection over Union ($\mIoU$), using MetaBox+ compared to when training with the full dataset, with only $10.47\%$ / $32.01\%$ annotation effort \commentMR{for the two networks, respectively.}
\end{abstract}

% \commentPC{Pascal}, \commentMR{Matthias}, \commentLRK{Lutz}, \commentHG{Hanno}


% ----------------------------------------------------------------------------------------------------------------
% ------------------------------------             INTRODUCTION             --------------------------------------
\section{Introduction} \label{sec:intro}

In recent years, semantic segmentation, the pixel-wise classification of the semantic content of images, has become a standard method to solve problems in image and scene understanding \cite{Unet, PSPNet, Deeplab, HRnet}. Examples of applications are autonomous driving and environment understanding \cite{PSPNet, Deeplab, HRnet}, biomedical analyses \cite{Unet} and further computer visions tasks. Deep convolution\commentMR{al} neural networks (CNN) are commonly used in semantic segmentation. In order to maximize the accuracy of a CNN, a large amount of annotated and varying data is required, since with an increasing number of samples the accuracy increases only \outPC{logarithmic} \commentPC{logarithmically} \cite{sun2017revisiting}. 
For instance in the field of autonomous driving, fully and precisely annotated street scenes require an enormous (and tiring) annotation effort. Also biomedical applications, in general domains that require expert knowledge for annotation, suffer from high annotation costs.
Hence, from multiple perspectives (annotation) cost reduction while maintaining model performance is highly desirable. 

One possible approach is \emph{active learning} (AL), which basically consists of alternatingly annotating data and training a model with the currently available annotation\commentPC{s}. The key component in this algorithm that can substantially leverage the learning process is the so called query or acquisition strategy. The ultimate goal is to label the data that leverages the model performance most while paying with as small labeling costs as possible. For an introduction to AL methods, see e.g.\ \cite{settles2009active}.

% PC: Diese Figure skizziert die AL Methode und wird zu Ende Sec 3 referenziert. Sie ist hier platziert, damit die erste Seite eine Figure hat!!!
\begin{figure}
\centerline{\includegraphics[width=.495\textwidth]{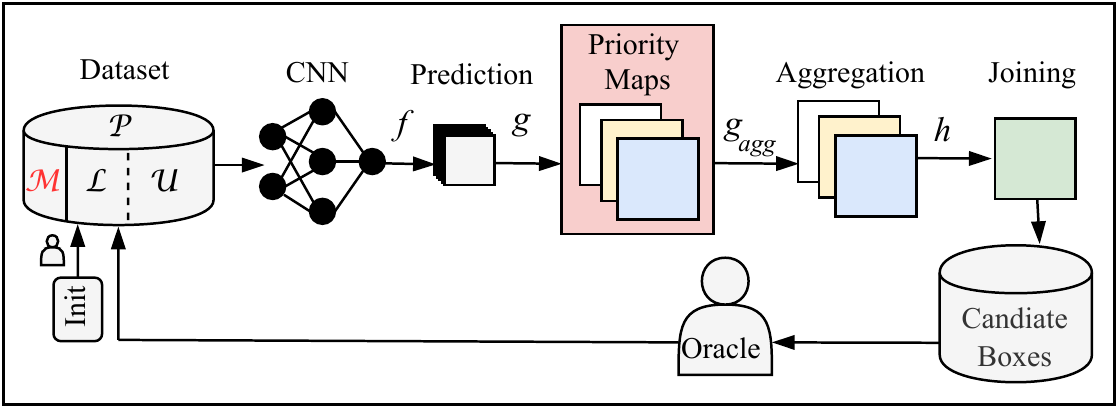}}
\caption{Illustration of the region based AL method. The red parts highlight the novel meta regression based ingredients: as query priority maps we use MetaSeg and the estimated number of clicks. Training of MetaSeg requires an additional small sample of data (fixed for the whole course of AL), indicated by $\mathcal{M} \subset \mathcal{P}$ (in red color).
} \label{fig:overall}
\end{figure}

% PC: Auf die Figure wurde bisher nicht referenziert. Ich habe die caption so geändert, dass man sich das Bild auch ohne Wissen der Methode(n) ansehen kann. Zusätzlich habe ich es in Sec 4, Paragraph "Comparison of MetaBox+ and EntropyBox+." referenziert, damit auch keine figure unreferenziert ist.
\begin{figure*}[]
\centerline{\includegraphics[width=0.99\textwidth]{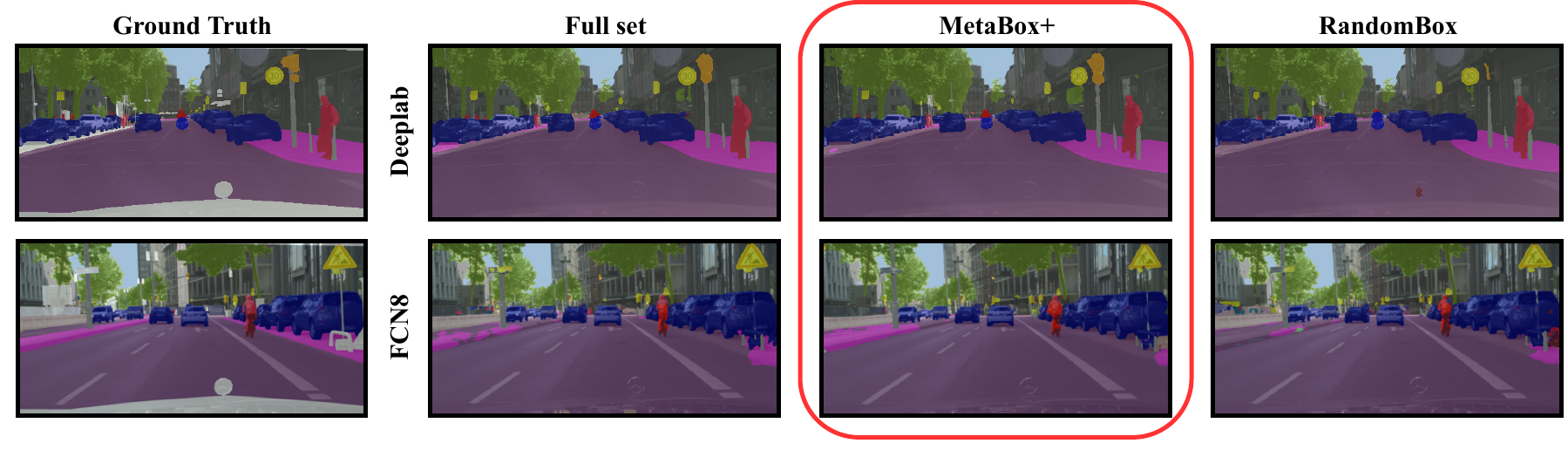}}
\caption{
% Segmentation results for both models: Deeplab  \emph{(top)} and FCN8  \emph{(bottom)}. The results for MetaBox+ and RandomBox are based on models with costs of $\cA = 10.47 \% $ and $\cA = 32.01 \%$ for Deeplab and FCN8 respectively, where MetaBox+ achieves $95 \% $  full set $\mIoU$.
\commentPC{
Segmentation results \outMR{with}\commentMR{with} our novel method MetaBox+ for two CNN models. With \outMR{only}an annotation effort of \commentMR{only} $10.47 \%$ for the Deeplab model \emph{(top)} and $32.01 \%$  for the FCN8 model \emph{(bottom)}, we achieve $95\%$ full set $\mIoU$.
Additionally, the segmentation results are shown when training the networks with the full dataset (Full set) and with a random selection strategy (RandomBox) producing the same annotation effort as MetaBox+. Annotation effort is \outMR{given by}\commentMR{stated in terms of} a click based metric ($\cA$, \Cref{eq:cA}).
}
}
\label{fig:segmentation}
\end{figure*}

First AL approaches (before the deep learning (DL) era) for semantic segmentation, for instance based on conditional random fields, go back to \cite{AL_SemSeg_expected_changes, AL_SemSeg_geometry, AL_SemSeg_curvlinear_structure, AL_SemSeg_active_propagation}.
At the heart of an AL method is the so-called \emph{query strategy} that decides which data to present to the annotator / oracle next.
In general, uncertainty sampling is one of the most common query \commentMR{strategies}\outMR{strategy} \cite{gal_deep_baysian_al, AL_pseudo_classification, AL_ensemble, hahn2019robustness, matthias2018deep}, besides that there also exist approaches on expected model change \cite{AL_SemSeg_expected_changes} and reinforcement learning based AL \cite{AL_SemSeg_Reinforced}. 
% An overview of query strategies can be found in \cite{settles2009active}.

In recent years, approaches to deep AL for semantic segmentation have been introduced, primarily for two applications, i.e., biomedical image segmentation and semantic street scene segmentation, cf.~\cite{AL_SemSeg_56_biomedical, AL_SemSeg_melanoma, AL_SemSeg_maxEntropy_medical, AL_SemSeg_GAN_mdeical, AL_SemSeg_ViewAL, AL_SemSeg_ReBased_Bosch, AL_SemSeg_Reinforced, CEREALS}.
%some work has also been developed in the field of AL for semantic segmentation using CNNs .
The approaches in \cite{AL_SemSeg_56_biomedical, AL_SemSeg_melanoma, AL_SemSeg_maxEntropy_medical, AL_SemSeg_GAN_mdeical} are specifically designed for medical and biomedical applications, mostly focusing on foreground-background segmentation. Due to the underlying nature of the data, these approaches refer to annotation costs in terms of the number of labeled images.
The methods presented in \cite{AL_SemSeg_ViewAL, AL_SemSeg_ReBased_Bosch, AL_SemSeg_Reinforced, CEREALS} use region based proposals.
All of them evaluate the model accuracy in terms of mean Intersection over Union ($\mIoU$). 
The method in \cite{AL_SemSeg_ViewAL} is designed for multi-view semantic segmentation datasets, in which \outMR{the}\outMR{class-}objects are observed from multiple viewpoints. The authors introduce two new uncertainty based metrics and aggregate them on superpixel (SP) level (SPs can be viewed as visually uniform clusters of pixels). They measure the costs by the number of labeled pixels. Furthermore they have shown that labeling on SP level can reduce the annotation time by $25 \%$.
In \cite{AL_SemSeg_ReBased_Bosch} a new uncertainty metric on SP level is defined, which includes the information of the Shannon Entropy \cite{shannon1948}, combined with information about the contours in the original image and a class-similarity metric to put emphasis on rare classes. In \cite{AL_SemSeg_ReBased_Bosch}, the number of pixels labeled define the costs.
The authors of \cite{AL_SemSeg_Reinforced} \outMR{perform}\commentMR{utilize} the same cost metric. \outMR{This paper}\commentMR{The latter work} uses reinforcement learning to find the most informative regions, which are given in quadratic format \outMR{with}\commentMR{of} fixed size. This procedure aims at finding regions containing instances of rare classes.
The method in \cite{CEREALS} queries quadratic regions of fixed-size from images as well. They use a combination of uncertainty measure (Vote~\cite{dagan1995committee} and Shannon Entropy) and a clicks-per-polygon based cost estimation, which is regressed by a second DL model.
The methods in \cite{CEREALS,  AL_SemSeg_Reinforced, AL_SemSeg_ReBased_Bosch} are focusing on multi-class semantic segmentation dataset\commentMR{s} like Cityscapes\outMR{.}~\cite{cityscapes}.

In this work, we introduce a query strategy, which is based on an estimation of the segmentation quality. We use a meta regression model, as introduced in \cite{MetaSeg} and extended in \cite{Schubert2019,Maag2019,Chan2019}, to predict the segment-wise $\IoU$ and then aggregate this information over the quadratic candidate \outMR{image regions}\commentMR{regions of images}. Furthermore, we introduce a simple and practical method to estimate the annotation costs in terms of clicks. Through the combination of both, we target informative regions with low annotation costs.
A sketch of our method is given in \Cref{fig:overall}.
Based on the number of clicks required for annotation, we introduce a new cost metric. For numerical experiments we used the Cityscapes dataset \cite{cityscapes} with two models, namely the FCN8~\cite{FCN8} and the Deeplabv3+ Xcpetion65~\cite{Deeplab} (following in short only Deeplab).

% This paper is structured as follows: in \outMR{section} \Cref{sec:work} we point out the differences between state-of-the-art region based AL methods for semantic segmentation and our method. Afterwards, in \outMR{section} \Cref{sec:metabox} we describe our AL method MetaBox+ in detail. In \Cref{sec:experiments}, we present and discuss our numerical results. Finally, \outMR{we give a} conclusion and outlook \commentMR{are given} in \outMR{section} \Cref{sec:conclusion}.

% ----------------------------------------------------------------------------------------------------------------
% ------------------------------------             RELATED WORK             --------------------------------------
\section{Related Work} \label{sec:work}

In this section, we compare our work \commentPC{to} the works closest to ours. Therefore, we focus on the region based approaches \cite{AL_SemSeg_ViewAL, AL_SemSeg_ReBased_Bosch, AL_SemSeg_Reinforced, CEREALS}. 
All of them evaluate the model accuracy in terms of $\mIoU$.
The approaches in \cite{AL_SemSeg_ReBased_Bosch, AL_SemSeg_ViewAL} use handcrafted uncertainty metrics aggregated on SP level and also query regions in form of SP.
While \cite{AL_SemSeg_ViewAL} is specifically designed for multi\commentPC{-}views datasets, our approach focus\commentMR{es} on single\commentPC{-}view multi-class segmentation of street scenes. 
For the query strategy presented in \cite{AL_SemSeg_ReBased_Bosch}, the authors do not only use uncertainty metrics, but also information \outMR{of}\commentMR{about} the contours of the images as well as class similarities of the SP to identify rare classes. We also use different types of information to \outMR{make}\commentMR{generate} our proposals. 
The AL method in \cite{AL_SemSeg_Reinforced} is based on deep reinforcement learning and queries quadratic fixed size regions. We query the same region format but instead of reinforcement learning, we only use the segmentation \commentMR{network's} output in our query strategy.
Compared to the approaches above \cite{AL_SemSeg_ReBased_Bosch, AL_SemSeg_ViewAL, AL_SemSeg_Reinforced} we do not focus on finding a minimal dataset to achieve a satisfying model accuracy. We consider the costs in terms of required clicks for annotating a region and aim to reduce the human annotation effort (therefore we refer to those clicks as costs). To this extent, we take an estimation of the labeling effort during acquisition into account. In addition, we estimate the segmentation quality to identify regions of interest.

As in \cite{CEREALS}, our candidate regions for acquisition are square-shaped and of fixed size, and \outPC{as in \cite{CEREALS}} we also use a cost metric based on the number of clicks required to draw a polygon overlay for an object. Hence, \cite{CEREALS} is in spirit closest to our work.
However, instead of using an uncertainty measure to identify high informational regions, we use information about the estimated segmentation quality in terms of the segment-wise Intersection over Union ($\IoU$) (see \cite{MetaSeg,Jaccard12similarityCoefficient}). Entropy is a common measure to \outMR{evaluate the}\commentMR{quantify} uncertainty. However, in semantic segmentation we observe increased uncertainty\outMR{values} on \outMR{the segments border}\commentMR{segment boundaries}, while \commentPC{uncertainty in}\outPC{\commentMR{uncertainties in}} the interior \outMR{values}\outPC{are}\commentPC{is} often low. This \outMR{observation}is in line with the \outMR{fact}\commentMR{observation} that, neural network in general provide overconfident predictions \cite{Goodfellow14Adversarial, hein2019relu}.
We solve this problem by evaluating the segmentation quality of whole predicted segments. 

Furthermore, also our cost estimation method differs substantially from the one presented in \cite{CEREALS}. While the authors of \cite{CEREALS} use another CNN to regress on the number of clicks per candidate region, we infer an estimate of the number of required clicks directly from the prediction on the segmentation network and show in our results, that our measure is indeed strongly correlated with the true number of clicks per candidate region.

% ----------------------------------------------------------------------------------------------------------------
% ------------------------------------             METHOD (MAIN)            --------------------------------------
\section{Region based Active Learning} \label{sec:metabox}

In this section we first describe a region based AL method, which queries fixed-size and quadratic image regions. 
Afterwards, we describe our new AL method. This method is subdivided into a 2-step process: first, we
predict the segmentation quality by using the segment\commentPC{-}wise meta regression method proposed in \cite{MetaSeg}. Second, we incorporate a cost estimation of the click number required to label a region, we term this add on.

% SUBSECTION -----------------------------------------------------------------------------------------------------
\subsection{Method Description} \label{sec:method_descr}

For the AL method we assume a CNN as semantic segmentation model with pixel-wise softmax probabilities as output\outMR{, called \emph{prediction}}. The \commentMR{corresponding} segmentation mask, also called \emph{segmentation}, is the \commentMR{pixel-wise} application of the $\argmax$ function to the \outMR{prediction}\commentMR{softmax probabilities}. 
The dataset is given as data pool $\mathcal{P}$. The set of all labeled data is denoted by
$\mathcal{L}$ and set of unlabeled data by $\mathcal{U}$.
At the beginning we have no labeled data, i.e., $\mathcal{L} = \emptyset$\commentMR{,}
and we wish to provide labels \outMR{for all images}over a given number of classes $c \in \mathbb{N}, c \geq 2$ \commentMR{for all images}.

A generic AL method \outMR{works}\commentMR{can be summarized} as follows: Initially, a small set of data from $\mathcal{U}$ is labeled and added to $\mathcal{L}$. Then, two steps are executed alternatingly in a loop. Firstly, the model (the CNN) is trained on $\mathcal{L}$. Thereafter, a chosen amount of unlabeled data from $\mathcal{U}$ is queried according to a query strategy, labeled and added to $\mathcal{L}$.

In region based AL, we \outMR{now}add newly labeled regions to $\mathcal{L}$ instead of whole images. An image $x$ remains in $\mathcal{U}$ as long as it is not entirely labeled. In order to avoid multiple queries of the same region, a region that is contained in both $\mathcal{U}$ and $\mathcal{L}$ is \outMR{equipped}\commentMR{tagged} with a \commentMR{query} priority equal to zero\outMR{for the query function}. In the remainder of this section, we describe the query function in detail and introduce an appropriate concept of priority in the given context.

\paragraph{Region based Queries.}
The \outPC{Query} \commentPC{query} strategy is a key ingredient of an AL method. \outMR{In region based methods we would like to have well informed queries, i.e., queries of which we expect that they leverage the training progress}\commentMR{In general, most query function designs strive for maximally leveraging training progress (i.e., achieving high validation accuracy after short time) at reduced labeling costs}. \commentMR{In the field of semantic segmentation, it seems}\outMR{As it might be} unnecessary to label whole images\commentMR{.} \commentMR{Thus,} we aim at querying regions of images \commentMR{which leads to a region-wise concept of query priority}. \outMR{Therefore, in our case \emph{priority} of an image region is a localized quantity.}

In what follows, we only compute \emph{measures of priority} \outPC{only} by means of the softmax output of the neural network. To this end, let 
\begin{equation}
    f : [0,1]^{w \times h \times 3} \to [0,1]^{w \times h \times c}
\end{equation}
be a function given by a segmentation network \commentMR{providing softmax probabilities for a given input image}, where $w$ denotes the image width, $h$ the height and $c$ the number of classes.

A \emph{priority map} can be viewed as another function
% $g: [0,1]^{w \times h \times c} \to [0,1]^{w \times h} $
\begin{equation} \label{eq:priority_heatmap}
    g: [0,1]^{w \times h \times c} \to [0,1]^{w \times h}
\end{equation}
that outputs \commentMR{one} priority score \commentMR{per} image pixel. The output of $g$ can be viewed as a heatmap that indicates priority.
A higher score of priority should presumably correlate with the attractiveness of the corresponding ground truth. A typical example for $g$ is the pixel-wise entropy $H$ which for a chosen pixel $(i,j)$ is given by
\begin{equation} \label{eq:entropy}
    H(y_{i,j,\cdot}) = - \sum_{k=1}^c y_{i,j,c} \log( y_{i,j,c} )
\end{equation}
where $y_{i,j,c} = f(x)_{i,j,c} \in [0,1]\outMR{^{w \times h \times c}}$ for a given input $x \in [0,1]^{w \times h \times 3}$. The priority maps that we use in our method are introduced in the subsequent section. Note that, if an image pixel has already been labeled, we overwrite the corresponding pixel value\outMR{s} of the priority map by zero\outMR{s}.

Our AL method queries regions that are square-shaped (\emph{boxes}) and of fixed \outMR{size}\commentMR{width} $b \in \mathbb{N}$. \outMR{For the sake of brevity, we use the shorthand \emph{boxes}.}A box-wise overall priority score is obtained via aggregation. To this end, we simply choose to sum up the scores. That is, given a box $B \subset [0,1]^{w \times h}$, the aggregated score is given by
\begin{equation}
    g_\mathit{agg}(y,B) =  \sum_{(i,j)\in B} g_{i,j}(y) \, .
\end{equation}

Given the set $\mathcal{B}$ of all possible boxes of \outMR{size}\commentMR{width} $b$ in $[0,1]^{w \times h}$, we can define an \emph{aggregated priority map}
\begin{equation} \label{eq:agg_inf_map}
    g_\mathcal{B}(y) = \{ g_\mathit{agg}(y,B) : B \in \mathcal{B} \}
\end{equation}
which can be viewed as another heatmap resulting from a convolution operation with a constant filter. \outMR{\commentPC{For the upcoming process, the output, named \emph{aggregated priority scores} has to be normalized with respect to the whole dataset, such that $g_\mathcal{B}(y) \subseteq[0,1]$. }}
Given $t$ aggregated priority scores, for the sake of brevity named $h^{(1)}(y,B),\ldots,h^{(t)}(y, B)$,
\outMR{\outMR{\commentPC{the} \outPC{a}} \commentMR{a} joint priority score\outMR{\commentPC{s}} can be defined \outMR{by} either \commentMR{by}}
we define a joint priority score by
% \begin{equation} \label{eq:pareto} % PC ohen Auskommentierung behält er das Label 
%     \outMR{h(y,B) = \sum_{s=1}^{t} \lambda_s h^{(s)}(y,B) \quad \text{with} \quad \sum_{s=1}^t \lambda_s = 1}
% \end{equation}
\outMR{or more lazily but parameter free}
\begin{equation} \label{eq:product}
    h(y,B) = \prod_{s=1}^{t} h^{(s)}(y,B) \, .
\end{equation}
\outMR{As AL is already computationally expensive, we avoid the Pareto optimization in XXX coming with additional parameters and focus on the latter approach.}
%\commentPC{Hier ist was gestrichen, aber Latex mag out Befehl mit Cref nicht (siehe TexCode.)}
\commentMR{Analogously to \Cref{eq:agg_inf_map} we introduce a joint priority map $h_\mathcal{B}(y)$.}
\commentMR{However, in} what follows we do not distinguish between joint priority maps and singleton (aggregated) priority maps as this follows from the context. Furthermore, we only refer to priority maps while performing calculations on \commentMR{priority score level}.
% \commentPC{\textbf{TODO: kurze Summary, da wir nur $g^{(i)}$, $h^{(i)}$ und $h$ benötigen.}}

\paragraph{Algorithm.} In summary, our AL method proceeds as follows. Initially, a randomly chosen set of $m_\mathit{init}$ entire images from $\mathcal{U}$ is labeled and then moved to $\mathcal{L}$. Afterwards, the AL \outMR{process is executed} \commentMR{method proceeds} as previously described in the introduction of this section. Defining the set of all \emph{candidate boxes} as
\begin{equation}
    C = \{ (y,B) : y=f(x), \, x \in \mathcal{U}, \, B \in \mathcal{B} \},
\end{equation}
we query in each iteration a chosen number $m_\mathit{q}$ of \emph{non-overlapping} boxes $Q = \{ (y_{i_j},B_j) : j=1,\ldots m_\mathit{q} \} \subset C $, with the highest \outMR{values}\commentMR{scores} $h(y,B)$, i.e.,
\begin{align}
    & (y,B) \in Q, (y',B') \notin \; Q \\
    \implies & h(y,B) \geq h(y, B')
    \text{ or } ( B \cap B' \neq \emptyset \text{ and } y=y' ) \, \nonumber .
        % B \in Q, B' \notin Q \implies g(B) \geq g(B') \, .
\end{align}
A sketch of the whole AL loop is depicted by \Cref{fig:overall}.

% SUBSECTION -----------------------------------------------------------------------------------------------------
\subsection{Joint priority maps based on meta regression and click estimation} \label{priority_metaseg_costs}

It remains to specify the priority maps $\outMR{h^{(i)} :=} h^{(i)} (y, B)$ defined in the previous section. In our method, we have $t=2$ priority maps. As an estimate of prediction quality, we use \emph{MetaSeg} \cite{MetaSeg} which provides a \outMR{separate}quality estimate in $[0,1]$ for each segment predicted by $f$. This aims at querying ground truth for image regions that presumably have been predicted badly. Mapping predicted qualities back to each pixel of a given segment and thereafter aggregating the values over boxes\commentMR{,} we obtain our first priority map $h^{(1)}$.

On the other hand, we wish to label regions that are easy (or cheap) to label. Therefore, we estimate the number of clicks required to annotate a box $B$.
From this, we define another priority map $h^{(2)}$ which contains high values for regions with low estimated numbers of clicks and vice versa (details follow in the upcoming paragraphs).
We query boxes according to the product of priorities, i.e.,
\begin{align} \label{eq:query}
     h(y,B)  & = h^{(1)}(y,B) \cdot h^{(2)}(y,B) %\\
\end{align}
as being done in \cite{CEREALS}, but with both $h^{(1)}$ and $h^{(2)}$ being different. In what follows\commentPC{,} we describe the priority maps $g^{(1)}(y)$ and $g^{(2)}(y)$ more precisely, where \outPC{are} the aggregated priority maps $h^{(1)}(y, B)$ and $h^{(2)}(y, B)$ \outMR{are based on \outMR{via} \commentMR{the} construction in}\commentMR{are constructed as in} \Cref{sec:method_descr}.

\paragraph{Priority via MetaSeg.}
As priority map $g^{(1)}(y)$ we use \emph{MetaSeg} \cite{MetaSeg} which estimates the segmentation quality by means of predicting the $\IoU$ of each predicted segment with the ground truth.

MetaSeg uses regression models with different types of hand-crafted input metrics. These include pixel-wise dispersion measures like (Shannon) entropy and the difference between the two largest softmax probabilities. These pixel-wise dispersions are aggregated on segment level by computing the mean over each segment. Here, a segment is a connected component of a \commentMR{predicted} segmentation mask of a given class.

In addition, for each predicted segment we consider shape-related quantities, i.e., the 
segment size, \commentMR{the fractality and the} surface center of mass coordinates\outMR{ and the predicted class}.
Furthermore, averaged class probabilities for each predicted segment are presented to the regression model.

Training the regression model of MetaSeg requires segmentation ground truth to compute the $\IoU$ for each predicted segment and the corresponding ground truth. 
Since the prediction changes in every iteration of the AL method, we train the regression model for MetaSeg once in every AL iteration.
In order to have ground truth available for training the regression model, we randomly select and label a further initial dataset $\mathcal{M}$ of $n_\mathit{meta}$ samples, which will be fixed for the whole AL process. 
To predict the quality of network predictions via MetaSeg, we perform the following steps after updating the semantic segmentation model:
\begin{enumerate}
    \item Infer the current CNN's predictions for all images in $\mathcal{M}$,
    \item Compute the metrics for each predicted segment (from step 1.),
    \item \textbf{Train MetaSeg} to predict the $\IoU$ by means of the metrics from step 2.
    \item Infer the current CNN's predictions for the unlabeled data $\mathcal{U}$,
    \item Compute the metrics for each predicted segment that belongs to $\mathcal{U}$ (as in step 2.),
    \item \textbf{Apply MetaSeg} in inference mode to each predicted segment from $\mathcal{U}$ (from step 4.) and its metrics (from step 5.) to predict the $\IoU$.
\end{enumerate}

\begin{figure}[t]
\centerline{\includegraphics[width=.495 \textwidth]{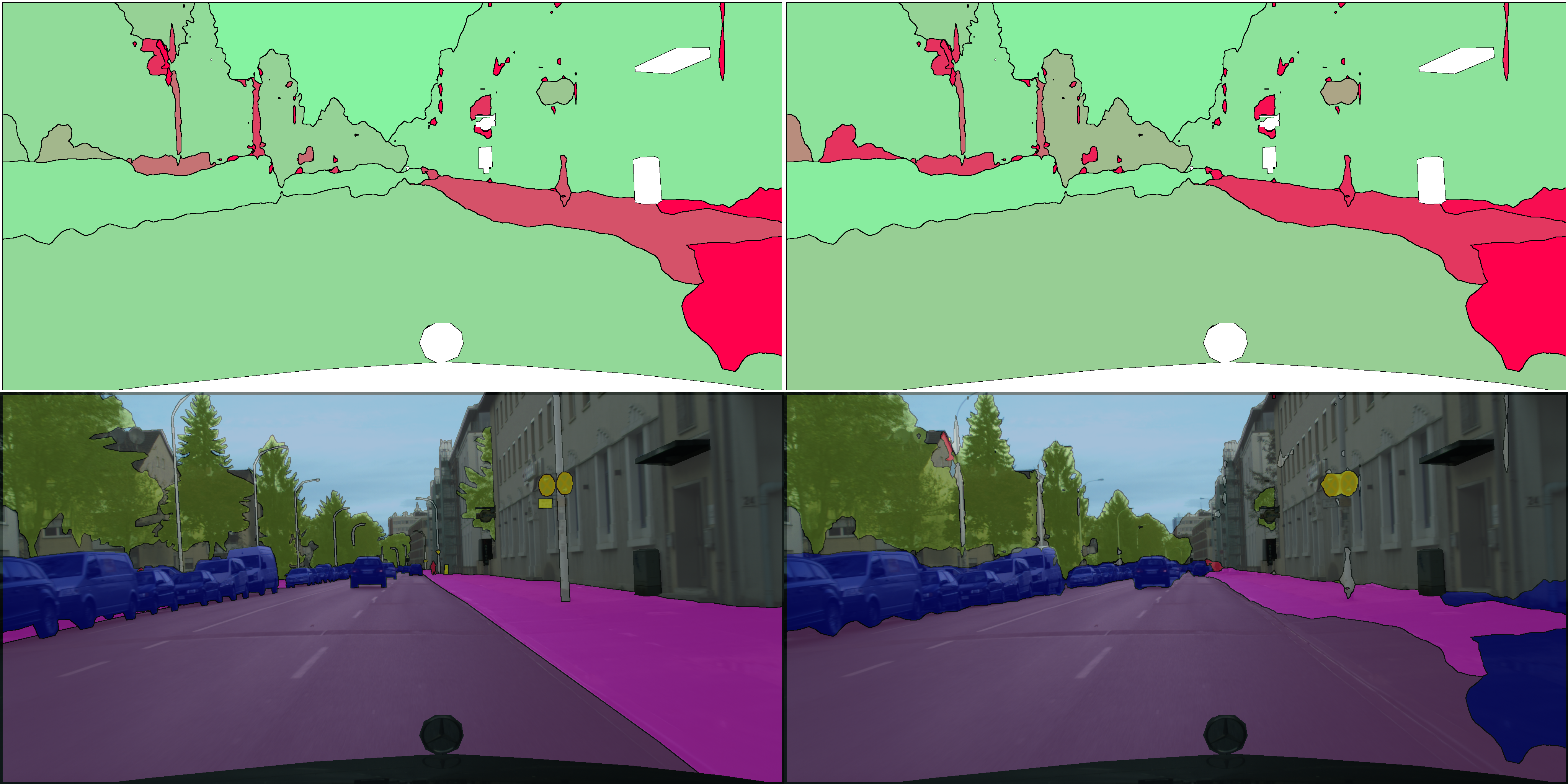}}
\caption{Prediction of the $\IoU$ values. The figure consists of ground truth  \emph{(bottom left)},
predicted segments \emph{(bottom right)}, true $\IoU$ for the predicted segments  \emph{(top left)} and predicted $\IoU$ for the predicted segments  \emph{(top
right)}. In the top row, green color corresponds to high $\IoU$ values and red color to low ones, for the white regions there is no ground
truth available.}
\label{fig:MetaSeg}
\end{figure}

For each unlabeled (i.e., not entirely labeled) image, MetaSeg provides a segmentation quality heatmap $q(y)$ by registering the predicted $\IoU$ values of the predicted segments for each of their corresponding pixels.
An example of the segmentation quality heatmap is given in \Cref{fig:MetaSeg}.
The corresponding priority map as defined in \Cref{eq:priority_heatmap}, is obtained via $g^{(1)}(y) = 1 - q(y)$. Hence, regions of $g^{(1)}(y)$ containing relatively high values are considered as being attractive for acquisition. More details on MetaSeg can be found in \cite{MetaSeg}.

\paragraph{Priority via Estimated Number of Clicks.} \label{cost_estimation}
As an additional priority map $g^{(2)}(y)$ we choose an estimate of annotation costs. 
Multi-class semantic segmentation datasets are generally labeled with a polygon based annotation tool, i.e., the objects are described by a finite number of vertices connected by edges such that the latter form a closed loop. \cite{cityscapes, neuhold2017mapillary}.
If the ground truth is given only pixel-wise, an estimate of the number of required clicks can be approximated by applying the Ramer-Douglas-Peucker (RDP) algorithm \cite{ramer1972iterative, douglas1973algorithms} to the segmentation contours. 

To estimate the true number of clicks required for annotation in the AL process, we correlate this number with how many clicks it approximately requires to annotate the predicted segmentation (provided by the current CNN) using the RDP algorithm. The approximation accuracy of the RDP algorithm is controlled by a parameter $\epsilon$.
% We use the implementation in OpenCV \cite{opencv_library} and set the approximation accuracy epsilon to $1.85$.
We define a cost map $\kappa(y) $ via $\kappa_{i,j}(y) = 1$ if there is a polygon vertex in pixel $(i,j)$ and $\kappa_{i,j}(y)= 0$ else.
Since we are prioritising regions with low estimated costs, the priority map is given by $g^{(2)}(y) = 1 - \kappa(y)$.
Following the construction in \Cref{sec:method_descr} yields the aggregated priority map $h^{(2)}(y, B)$ \outPC{ for candidate boxes $(y,B) \in C$}.
A \commentPC{visual} example \commentPC{of the cost estimation} is given in \Cref{fig:click_img_annonationg} (left panel).

In our tests with the RDP algorithm applied to ground truth segmentations, we observed that on average the estimated number of click is fairly close to the true number of clicks provided by the \commentPC{Cityscapes} \outPC{cityscapes} dataset.
Therefore, if we assume that over the course of AL iterations, the model performance increases, approaching a level of segmentation quality that is close to ground truth, then the described cost estimation on average will approach the click numbers in the ground truth.
\begin{figure*}[t]
\centerline{\includegraphics[width=0.99 \textwidth]{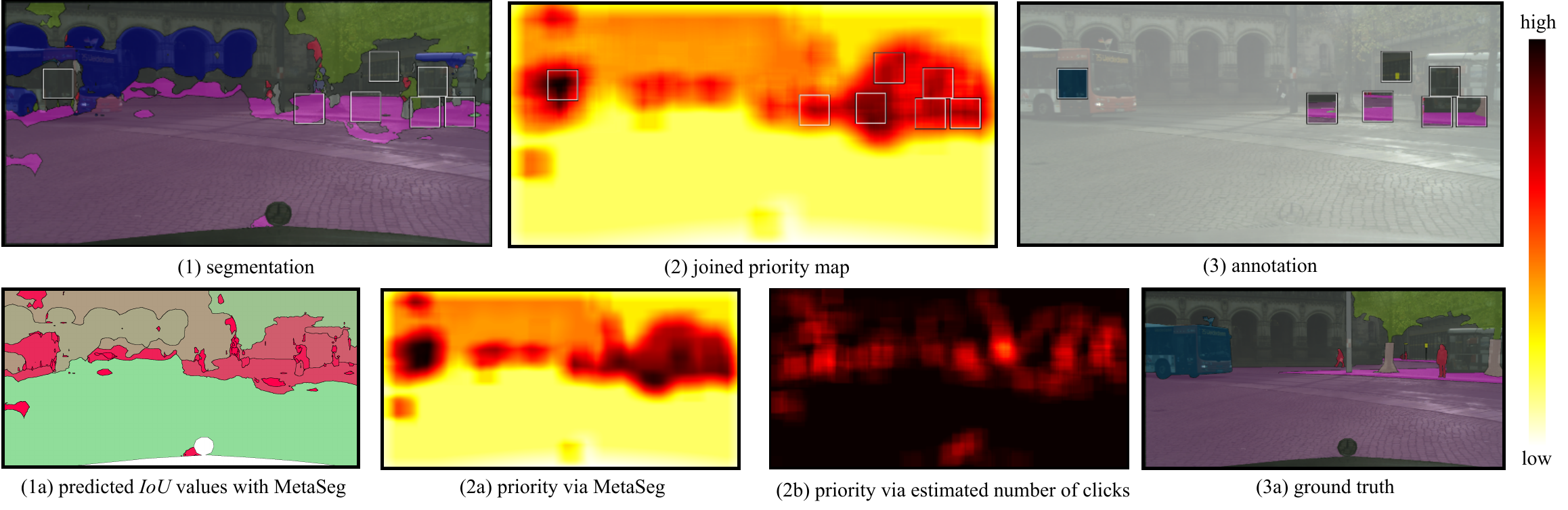}}
\caption{Visualisation of our AL method MetaBox+ at a specific AL iteration. The top row shows the segmentation (1), the joined priority map (2) and the acquired annotation (3).
The joined priority map (2) is based on the priority via MetaSeg (2a) and the estimated number of clicks (2b). High values represent prioritized regions for labeling: in (2b) regions with low predicted $\IoU$ values are of interest in (2b) regions with low estimated clicks. (1a) shows the predicted $\IoU$ values via MetaSeg.
}
\label{fig:method_vis}
\end{figure*}

In the following, we distinguish\outMR{ed} between the \commentMR{following} methods: \emph{MetaBox} uses only the priority via MetaSeg and \emph{MetaBox+} \outMR{\outPC{uses} \commentPC{using}}\commentMR{uses both} the joint priority of MetaSeg and the estimated number of clicks.
\outMR{Our}\commentMR{An overview of the different steps of our} AL method is \commentMR{given by} \Cref{fig:overall} and a\commentMR{n exemplary} visualization of the different stages of MetaBox+ is shown in \Cref{fig:method_vis}. Note that\outPC{,} there are different conventions for counting clicks which we discuss in \Cref{sec:annotationeffort}.

\paragraph{Further priority maps \commentMR{and baseline methods}.}
For the sake of comparison, we also define a priority map based on the pixel-wise entropy as in \Cref{eq:entropy}. Analogously to MetaBox and MetaBox+, we introduce \emph{EntropyBox} and \emph{EntropyBox+}: EntropyBox uses only the priority via entropy and EntropyBox+ uses the joint priority of the entropy and the estimated number of clicks. The method EntropyBox+ is similar to the method introduced in \cite{CEREALS}. The corresponding authors also use a combination of the entropy and a cost estimation, but the cost estimation is computed by a second DL model. 
Furthermore, as a naive baseline we consider a random query function that performs queries by means of random priority maps. We term this method \emph{RandomBox}.

% ----------------------------------------------------------------------------------------------------------------
% ------------------------------------             EXPERIMENTS (MAIN)       --------------------------------------
\section{Experiments}\label{sec:experiments}

\commentMR{Before presenting results of our experiments} \outMR{In this section}, we introduce metrics to measure the annotation effort. To this end, we discuss different types of clicks required for labeling and how they can be taken into account for defining annotation costs. \commentPC{Afterwards we specify the experiments settings and the implementation details}. Thereafter, we present numerical experiments where we compare different query strategies with respect to performance and robustness. Furthermore we study the impact of incorporating annotation cost estimations.

% SUBSECTION -----------------------------------------------------------------------------------------------------
\subsection{Measuring Annotation Effort} \label{sec:annotationeffort}

In semantic segmentation, annotation is usually generated with a polygon based annotation tool. A connected component of a given class is therefore represented by a polygon, i.e., a closed path of edges. This path is constructed by a human labeler clicking at the corresponding vertices. We term these vertices \emph{polygon clicks} $c_{p}(B) \in \mathbb{N}_0$.
Since we query quadratic image regions (boxes), we introduce the following additional types of clicks:
\begin{itemize}
    \item \emph{intersection clicks}, $c_{i}(B) \in \mathbb{N}_0$, occur due to the intersection between the contours of a segment and the box boundary,
    \item \emph{box clicks}, $c_{b}(B) \in \mathbb{N}_0$, specify the quadratic box itself,
    \item \emph{class clicks}, $c_{c}(B) \in \mathbb{N}$, specify the class of the annotated segment.
\end{itemize}
\commentPC{For an annotated image, the class clicks correspond to the number of segments. Like the polygon clicks, they can also be considered for the cost evaluation of fully annotated images.}

For the evaluation of a dataset $\mathcal{P}$ (with fully labeled images), $c_{p}(\mathcal{P}) \in \mathbb{N}$ is the total number of polygon clicks and $c_{c}(\mathcal{P}) \in \mathbb{N}$ the total number of class clicks.
Let $\mathcal{L}_0$ be the the initially annotated dataset (with fully labeled images) and $Q$ the set of all queried and annotated boxes, then we define the cost metrics
\begin{align}
    & \cA = \frac{ c_p(\mathcal{L}_0) + c_c(\mathcal{L}_0) +  c_p(Q) + c_i(Q) + c_c(Q)}{ c_{p}(\mathcal{P})  + c_{c}(\mathcal{P})} \label{eq:cA} \\
    & \cB = \frac{ c_p(\mathcal{L}_0) +  c_p(Q) + c_i(Q) + c_b(Q)}{ c_{p}(\mathcal{P})} \label{eq:cB} \\
    & \text{ with } c_{\sharp} (Q) = \sum\limits_{ B_j \in Q } c_{\sharp} (B_j), \quad \sharp\in \{p,i,b,c\} \nonumber \text{.}
\end{align}
\commentPC{In addition}\outPC{Addition} to that,
\begin{align}
    \cP \label{eq:cP} 
\end{align}
 defines the costs as amount of labeled pixels with respect to the whole dataset.

The amount of required clicks depends on the annotation tool. The box clicks $c_{b}(B)$ are not necessarily required: with a suitable tool, the chosen image regions (boxes) are suggested and the annotation process restricted accordingly.
Required are the polygon clicks $c_{p}(B)$ and the intersections clicks $c_{i}(B)$ to define the segment contours as well as the class clicks $c_{c}(B)$ to define the class of the annotated segment. Cost metric $\cA$ (\Cref{eq:cA}) is based on this consideration.  

\begin{figure}
\centerline{\includegraphics[width=.495 \textwidth]{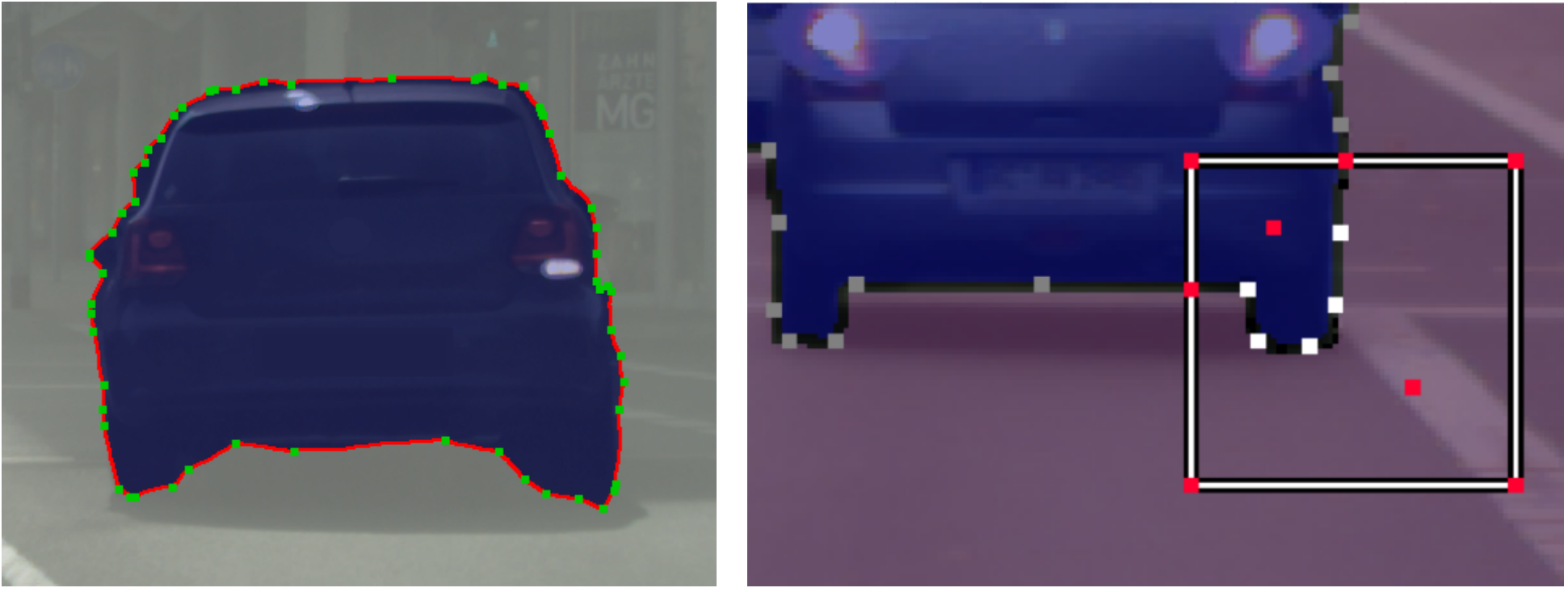}}
\caption{\emph{(left)}: Visualization of the estimated annotation clicks obtained by the RDP algorithm applied to the segmentation contours of a predicted segment of class ``car''. The segment contours are highlighted by red color, the obtained vertices (estimated clicks) are highlighted green. 
\emph{(right)}: Example of possible types of clicks we can take into account for a cost definition. The white (and gray) pixels depict true annotation clicks obtained from the data of (here $5$ polygon clicks within the box). The red ones represent additional types of clicks: the ones required for annotating intersection points of segment contour and box edges (here $2$ intersection clicks), the ones for defining the box itself ($4$ box clicks, one for each corner) and one click per segment to specify the class of the segment (here $2$ class clicks: for car and street).
} \label{fig:click_img_annonationg}
\end{figure}

Cost metric $\mathit{cost}_B$ (\Cref{eq:cB}) is introduced in \cite{CEREALS}. 
Due to a personal correspondence with the authors we are able to state details that go beyond the description provided in \cite{CEREALS}:
Cost metric $\cB$ is mostly in accordance with $\cA$, except for two changes.
The box clicks $c_{b}(B)=4$ are taken into account while the class clicks $c_{c}(B)$ are omitted. 
Theoretically both metrics can become greater than $1$. Firstly, fully labeled images do not require intersection clicks $c_{i}(B)$. Secondly, ground truth segments that are labeled by more than one box produce multiple class clicks $c_{c}$ to specify the class. \commentPC{In the following, the cost metrics $\cA, \ \cB, \ \cP$ are given in \commentMR{percent} \commentMR{(of the costs for labeling the full dataset without considering regions)}.}
An illustration of the click types is shown in \Cref{fig:click_img_annonationg} (right panel).

% monchengladbach_000000_012376

% SUBSECTION -----------------------------------------------------------------------------------------------------
\subsection{Experiment Settings}
For our experiments, we used the Cityscapes \cite{cityscapes} dataset. It contains images of urban street scenes with $19$ classes for the task of semantic segmentation. Furthermore, the annotation clicks / polygons are given. We used the training set with $2  ,\! 975$ samples as data pool $\mathcal{P}$. For all model and experiment evaluations we used the validation set containing $500$ samples. 
We used two CNN models: \emph{FCN8} \cite{FCN8} (with width multiplier $0.25$ introduced in \cite{mobilenet}) and Deeplabv3+ \cite{Deeplab} with an Xception65 \cite{xception} backbone, (short: \emph{Deeplab}).
Using all training data, also referred to as full set, we achieve a $\mIoU$ of $60.50 \%$ on the validation dataset with the FCN8 model and a $\mIoU$ of $76.11 \%$ for the Deeplab model. We have not resized the images, i.e., we used the original resolution of height $h=1 ,\! 024$ and width $w=2 ,\! 048$.
In each AL iteration, we train the model from scratch. The training \outMR{converges}\commentMR{is stopped}, if no improvement in term of validation $\mIoU$ is achieved over $10$ consecutive epochs. Details regarding the training parameters are given in \outMR{the}\Cref{sec:implementation_details} below.
All experiments started from an initial dataset of $50$ samples. For experiments with MetaSeg based queries (MetaBox, MetaBox+), we took $30$ additional samples to train MetaSeg. In each AL iteration we queried $6 ,\! 400$ boxes with a \outMR{size}\commentMR{width} of $b = 128$, which corresponds to $50$ full images in terms of the number of pixels.

% resources 
Each experiment was repeated three times. For each method we present the mean over the $\mIoU$ and the \commentPC{mean over the} cost metrics ($\cA$, $\cB$, $\cP$) of each AL iteration. All CNN \outPC{training's} \commentPC{trainings} were performed on NVIDIA Quadro P6000 GPUs.
In total, we trained the Deeplab model $180$ times, which required approximately $8,\!000$ GPU hours and the FCN8 model $290$ times, which required approximately $3,\!500$ GPU hours. This amounts to $11,\!500$ GPU hours in total. On top of that, we consumed a few additional GPU hours for the inference as well as a moderate amount of CPU hours for the query process.

% SUBSECTION -----------------------------------------------------------------------------------------------------
\subsection{Implementation Details} \label{sec:implementation_details}

To train the CNN models with only parts of the images, we set the labels of the unlabeled regions to ignore. 
Both models (FCN8 and Deeplab) are initialized with pretrained weights of imagenet \cite{ImageNet}.

\paragraph{FCN8.}
For training of the FCN8 model~\cite{FCN8}, with the width multiplier $0.25$ \cite{mobilenet}, we used the Adam optimizer
with learning rate, alpha, and beta set to $0.0001$, $0.99$, and $0.999$, respectively.
% Last_comment_PC: wir haben 3 Parameter (learning rate, alpha, beta set) mit den zugehörigen 3 werten. 
% with learning rate, alpha and beta set to $0.0001$, $0.99$, and $0.999$ \commentMR{\textbf{COMMENT:2 parameter mit 3 Werten? Vielleicht doch nicht respectively?}}, respectively.
We used a batch size of $1$ and did not use any data augmentation\commentPC{.} \outPC{since this is not part of the provided code.}

\paragraph{Deeplab.}
For the training of the Deeplab model, with the Xception-backbone \cite{xception} we proceeded as in \cite{Deeplab}: we set decoder output stride to $4$, train crop size to $769 \times 769$, atrous rate to $6, 12, 18$ and output stride to $16$. 
To consume less GPU memory resources we used a batch size of $4$: We have not fine-tuned the batch norm parameters. For the training in the AL iterations, we used as polynomial decay learning rate policy: 
\begin{equation*}
    % current \ lr = base \ lr * ( \frac{1 - global \ step}{tot \ number \ steps} ) ^{learning \ power}
    \mathit{lr}^{(i)} = \mathit{lr}_{\mathit{base}} * \Big( \frac{1 - s^{(i)}}{s_{\mathit{tot}}}\Big)^p
\end{equation*}
where $\mathit{lr}^{(i)}$ is the learning rate in step $s^{(i)} = i $, $lr_{\mathit{base}} = 0.001$ the base learning rate, $p=0.8$ the learning power and $s_{\mathit{tot}} = 150,\!000$ the total number of steps.
For training with the full set we used the same learning rate policy with a base learning rate $\mathit{lr}_{base} = 0.003$.
With these settings we achieve a $\mIoU$ of $76.11 \% $ (mean of $5$ runs). The original model achieves a $\mIoU$ of $78.79 \% $ \commentMR{(with a batch size of 8)}.

\paragraph{MetaSeg.} 
We used the implementation of \url{https://github.com/mrottmann/MetaSeg} with minor modifications in the regression model.
Instead of a linear regression model, we used a gradient boosting method with $100$ estimators, max depth $4$ and learning rate $0.1$. \commentPC{\commentMR{In our tests,} a gradient boosting method \commentMR{led} to better results than a linear regression model.} For the training of MetaSeg we used $30$ images. We tested MetaSeg for different number\commentMR{s} of predictions and of different\commentMR{ly} performing CNN models. With the given parameters, \commentMR{we achieve results in terms of $R^2$ values similar to those} presented in the original paper \cite{MetaSeg}.

% SUBSECTION -----------------------------------------------------------------------------------------------------
\subsection{Evaluation}

% PC: Hinzunahme von RandomBox im Paragraphen, da die Erkenntnisse ebenfalls aufgeführt werdne und wichitg sind
% \paragraph{Comparison of MetaBox and EntropyBox.}
\paragraph{\commentPC{Comparison of MetaBox, EntropyBox and RandomBox.}}

\begin{figure}[h]
\centerline{
\includegraphics[width=.248\textwidth]{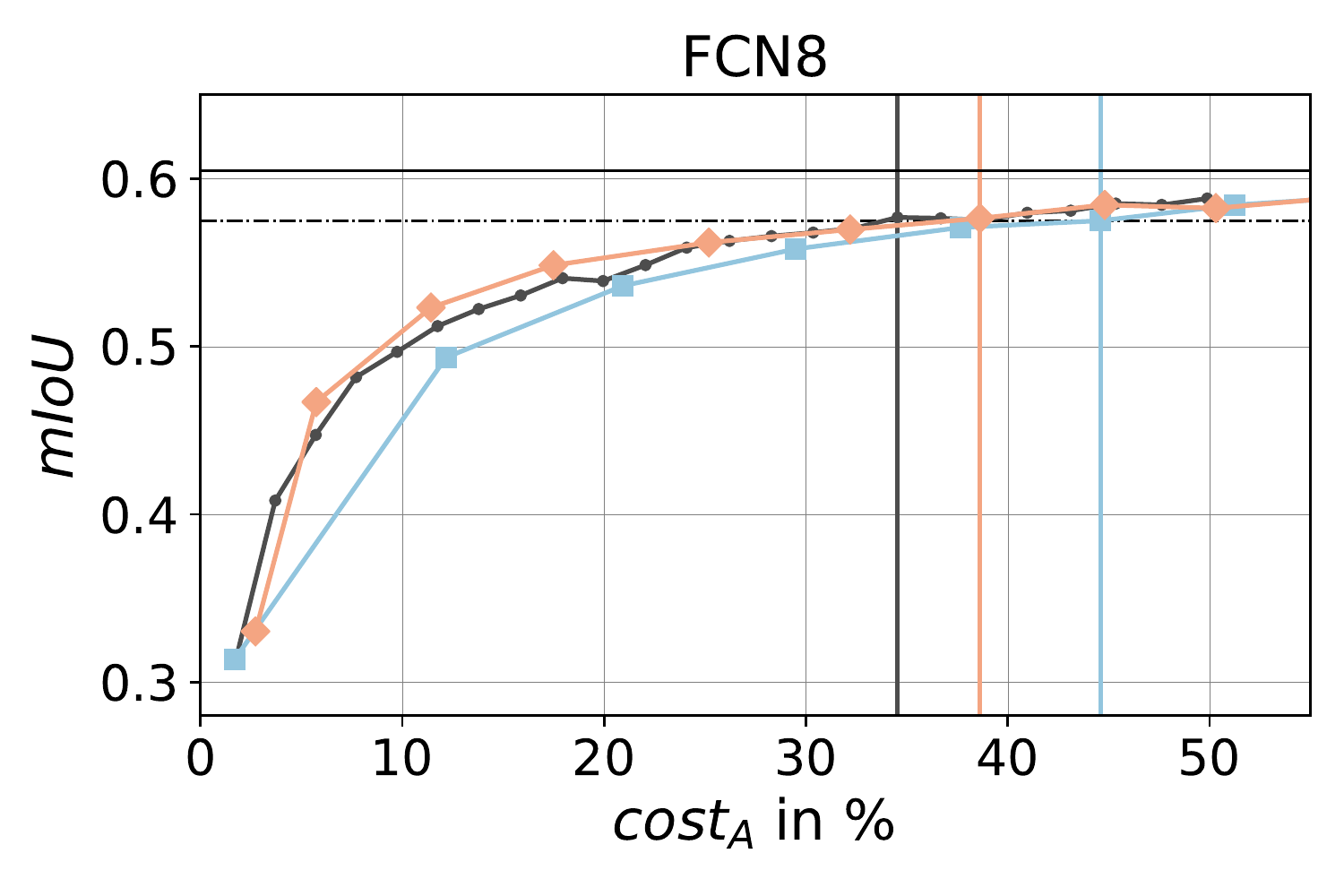} \hfill \includegraphics[width=.248\textwidth]{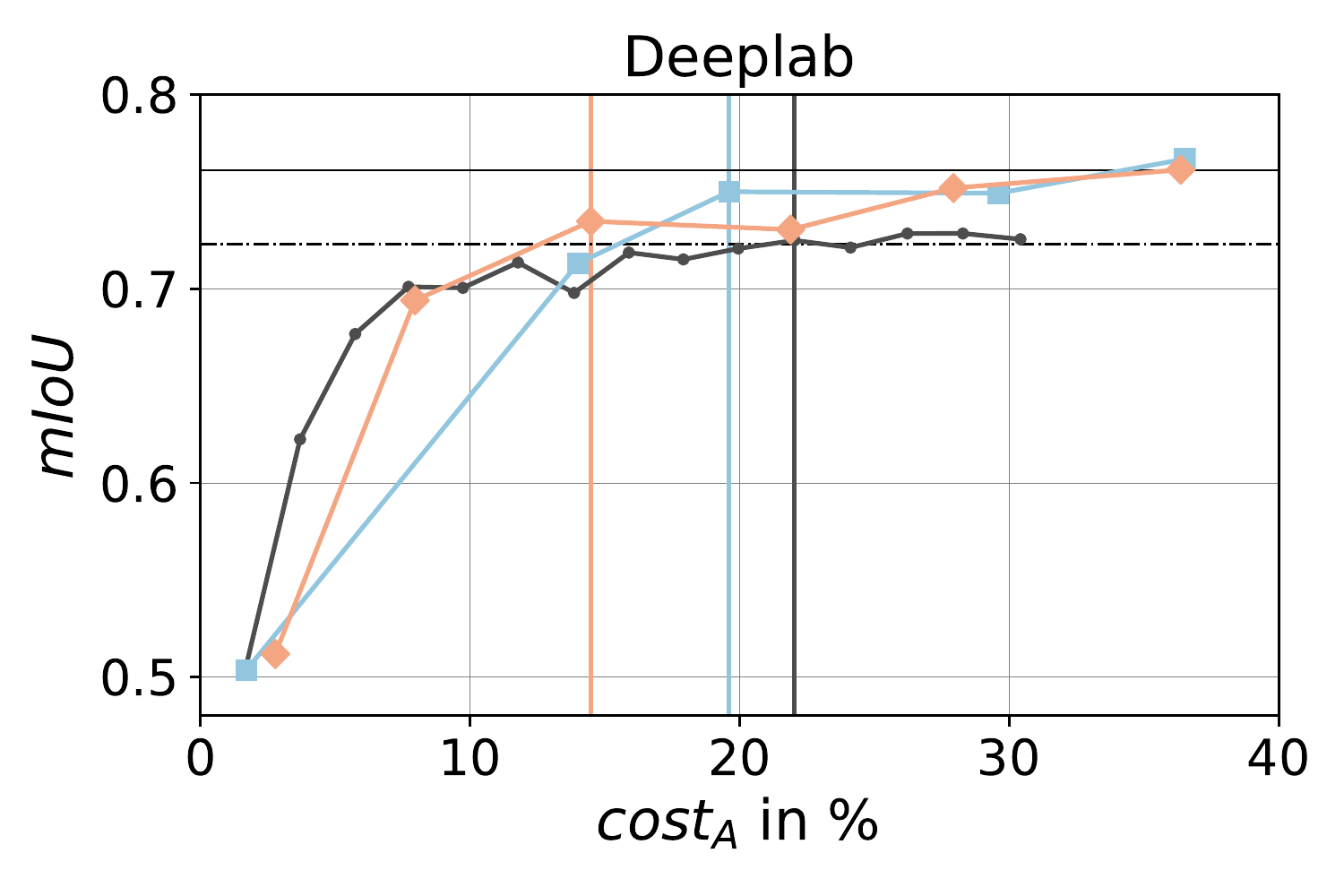}
}
\centerline{\includegraphics[width=.495\textwidth]{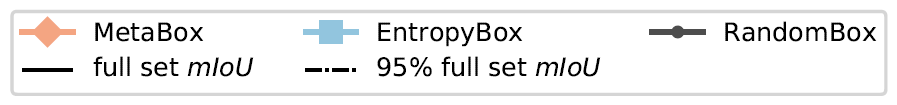}}
\caption{Results of the AL experiments for MetaBox, EntropyBox and RandomBox. Costs are given in terms of cost metric $\cA$. The vertical lines indicate where a corresponding method achieves $95\%$  full set $\mIoU$. Each method's curve represents the mean over 3 runs.}
\label{fig:experiments_no_cost_estimation}
\end{figure}

% PC: Wiedergabe der numerischen Ergebnisse 
First we compare the methods \outPC{MetaBox and EntropyBox}\outMR{,}\outMR{which}\commentMR{that} do not include the cost estimation. 
As \outPC{it}can be seen in \Cref{fig:experiments_no_cost_estimation}, MetaBox outperforms EntropyBox in terms annotation required to achieving $95 \%$  full set $\mIoU$: for the FCN8 model, MetaBox produces click costs of $\cA = 38.63 \%$ while EntropyBox produces costs of $\cA = 44.60 \% $. For analogous experiments with the Deeplab model, MetaBox produces click costs of $\cA = 14.48 \% $ while EntropyBox produces costs of $\cA = 19.61 \%$. Furthermore, for the Deeplab model both methods perform better compared to RandomBox, which requires costs of $\cA = 22.04 \%$. However, for the FCN8 model RandomBox produces the least click costs of $\cA= 34.54 \%$.
Beyond the $95 \%$ (full set $\mIoU$) frontier, all three methods perform very similar on the FCN8 model. For the Deeplab model, RandomBox does not significantly gain performance while MetaBox and EntropyBox achieve the full set $\mIoU$ requiring approximately the same costs of $\cA \approx 36.56 \%$.

\begin{figure}
\centerline{\includegraphics[width=.495\textwidth]{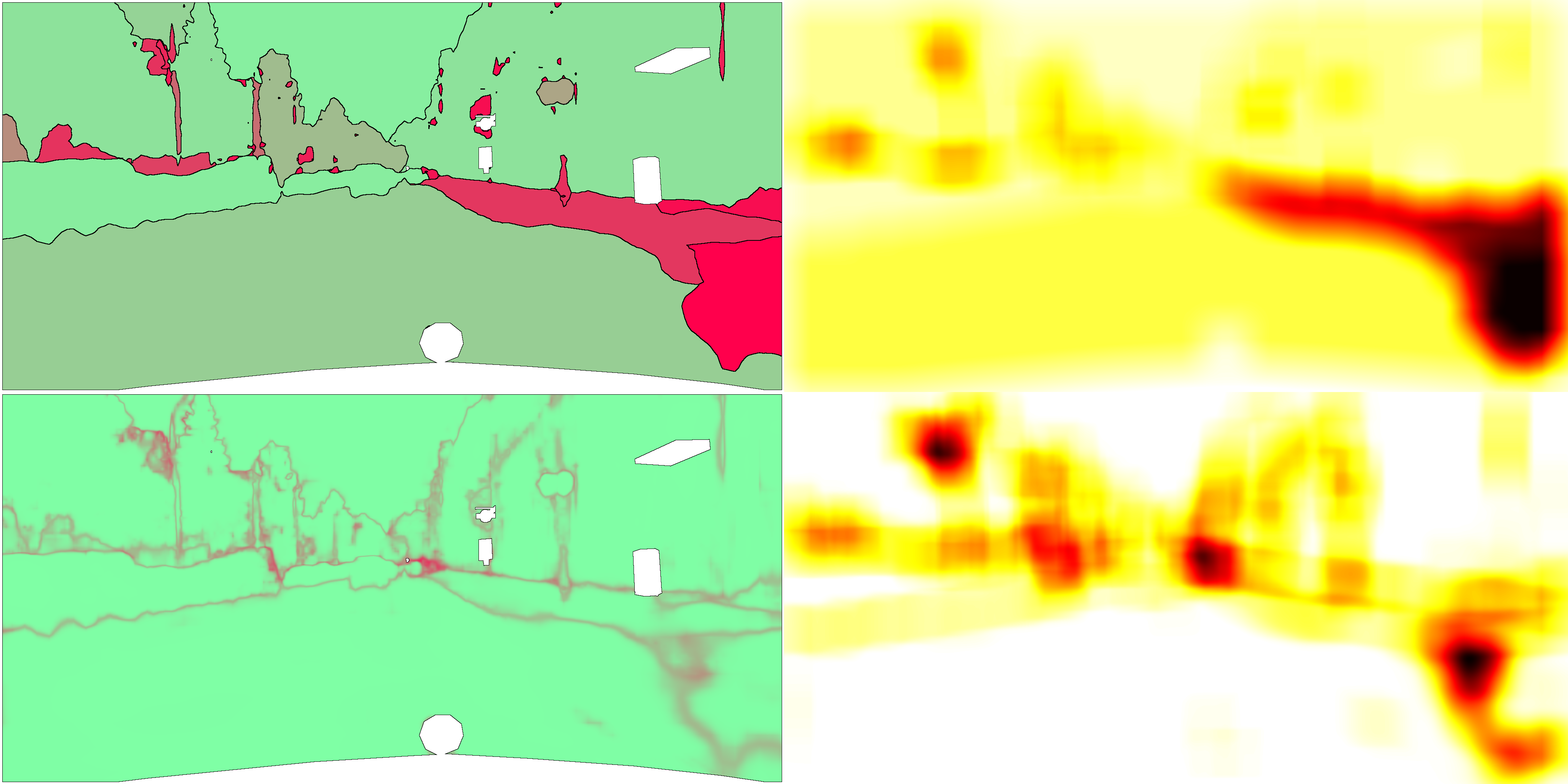}}
\caption{Comparison of query strategies based on MetaSeg and entropy. In the MetaSeg priority map  \emph{(top left)}, low estimated $\IoU$ values are colored red and high ones green. Accordingly, in the entropy priority map  \emph{(bottom left)} low confidence is colored red and high confidence is colored green. The (line-wise) corresponding aggregations are given in the right hand column. The higher the priority, the darker the color.
} \label{fig:Entropy_vs_MetaSeg}
\end{figure}

% PC: Erläuterung der Ergbenisse
\commentPC{\Cref{fig:Entropy_vs_MetaSeg} shows a visualisation of prioritised regions for annotation. In general, high entropy values \commentMR{are observed on the boundaries} \outMR{correlate with the border}of predicted segments. Therefore EntropyBox queries boxes, which overlap with the contours of predicted segments. Since MetaBox prioritises regions with low predicted $\IoU$ values, queried boxes \outMR{mostly}\commentMR{often} lie in the interior of predicted segments. Furthermore, EntropyBox produces higher costs in each AL iteration compared to MetaBox and RandomBox. 
RandomBox produces relatively small but very consistent costs per AL iteration.
\outMR{An AL iteration with high costs is also associated with high improvement in the performance.}
}

\paragraph{Comparison of MetaBox+ and EntropyBox+.}

\begin{figure}
\centerline{
\includegraphics[width=.248\textwidth]{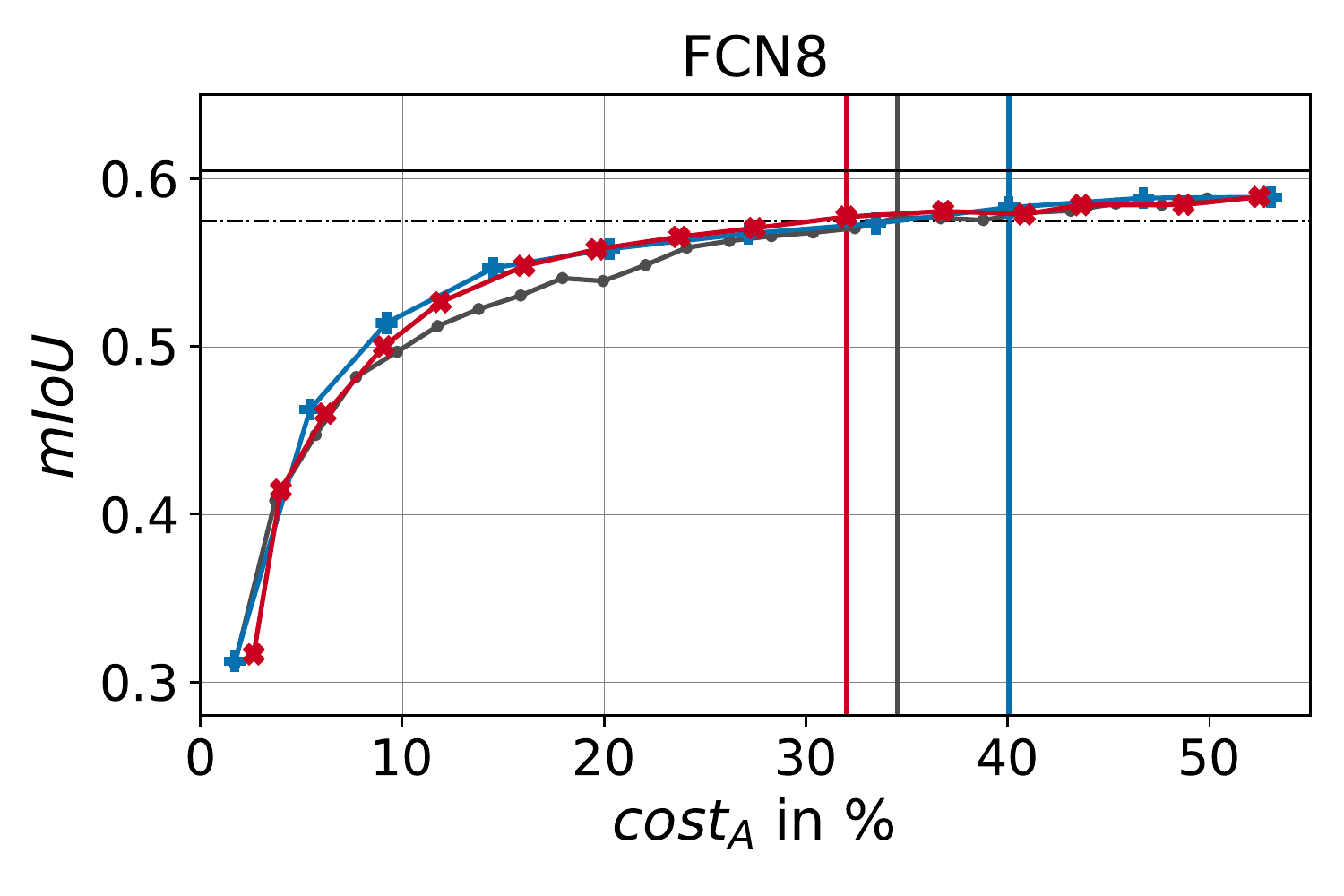} \hfill \includegraphics[width=.248\textwidth]{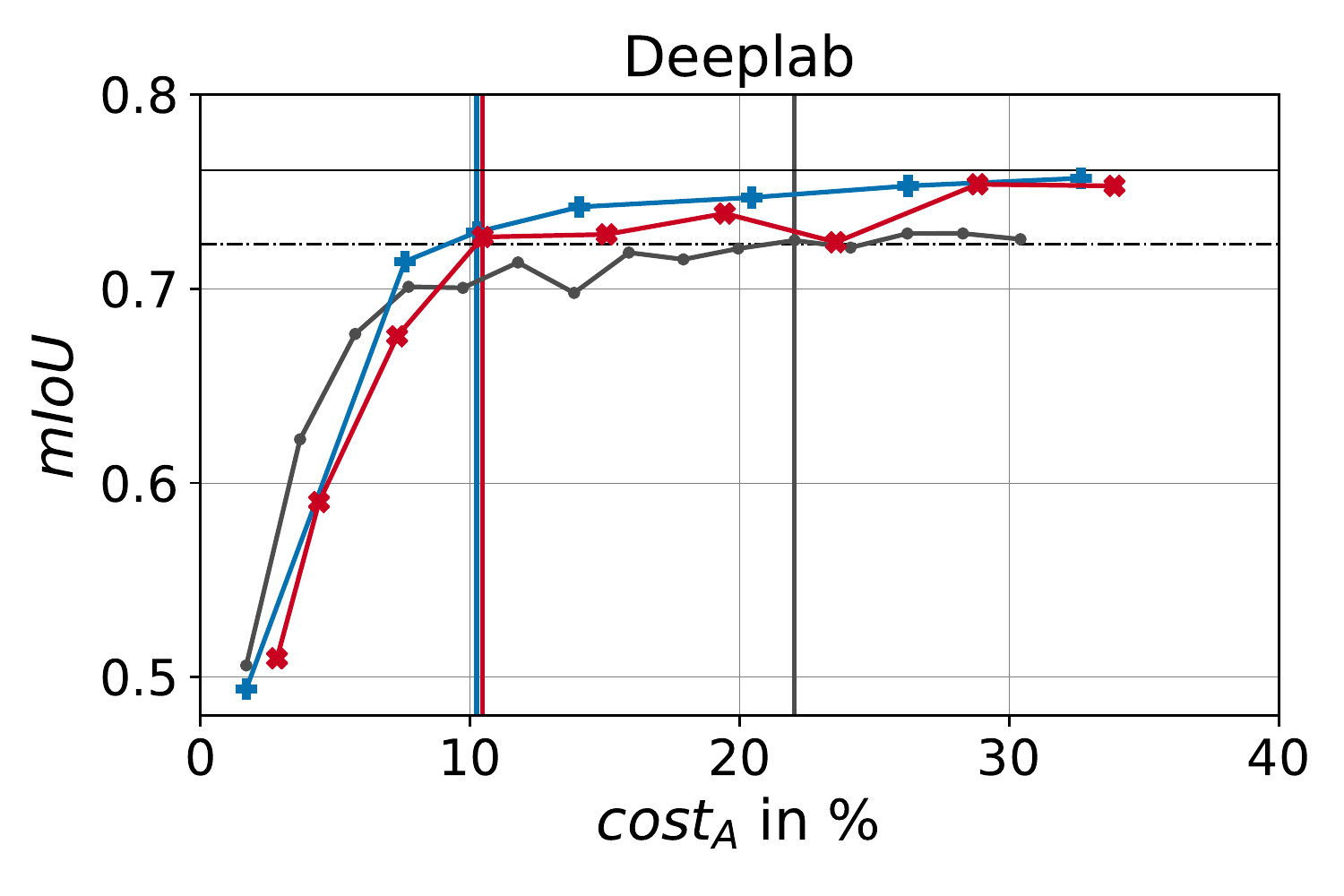}
}
\centerline{\includegraphics[width=.495\textwidth]{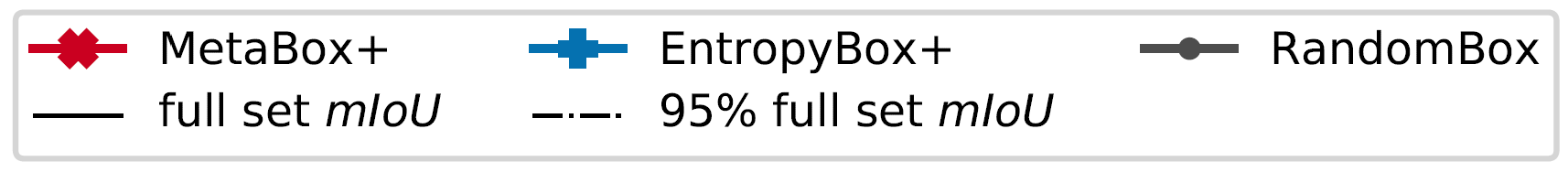}}

\caption{Results of the AL experiments for MetaBox+ and EntropyBox+. Costs are given in terms of cost metric $\cA$. The vertical lines indicate where a corresponding method achieves $95\%$  full set $\mIoU$. Each method's curve represents the mean over 3 runs.}
\label{fig:experiments_with_cost_estimation}
\end{figure}

Incorporating the estimated number of clicks improves both methods MetaBox and EntropyBox, see \Cref{fig:experiments_with_cost_estimation}. For the FCN8 model, EntropyBox+ still produces more clicks $\cA = 40.06 \%$ compared to RandomBox. On the other hand, MetaBox+ requires the \outMR{least}\commentMR{lowest} costs with $\cA = 32.01 \% $.
For the Deeplab model, EntropyBox+ and MetaBox+ produce almost the same click costs ($\cA = 10. 25 \%$ and $\cA = 10.47 \%$, respectively) for achieving $95 \%$  full set $\mIoU$. \commentPC{By taking the estimated costs into account, the produced costs per AL iteration are lower for both methods.}
% Last_comment_PC  Compared to EntropyBox and MetaBox
\commentPC{Although, in comparison with EntropyBox and MetaBox, the methods EntropyBox+ and MetaBox+ require more AL iterations, both methods perform better in terms of required clicks to achieve $95 \% $ full set  $\mIoU$.} 
% Qualitative segmentation results using MetaBox+ are shown in \Cref{fig:segmentation}.} }
% \outMR{Even if} \commentMR{Although} EntropyBox+ and MetaBox+ require more AL iterations \commentMR{\textbf{COMMENT: more than what?}}, both methods perform better in terms of required clicks to achieve $95 \% $ full set  $\mIoU$. \commentMR{\textbf{COMMENT: better than what?}}}\commentPC{Qualitative segmentation results using MetaBox+ are shown in \Cref{fig:segmentation}.}

In general, we observe that the Deeplab model gains performance quicker than the FCN8 model. This can be attributed to the fact that the FCN8 framework does not incorporate any data augmentation while the Deeplab framework uses state-of-the-art data augmentation and provides a more elaborate network architecture.

\begin{figure*}[h] % t for on top 
% \centerline{\includegraphics[width=.495\textwidth]{figs/fcn8/All_Experiments_cost_A_fcn8_95_no_var.pdf}} \centerline{\includegraphics[width=.495\textwidth]{figs/deeplab/All_Experiments_cost_A_deeplab_95_no_var.pdf}}
\centerline{
\includegraphics[width=.495\textwidth]{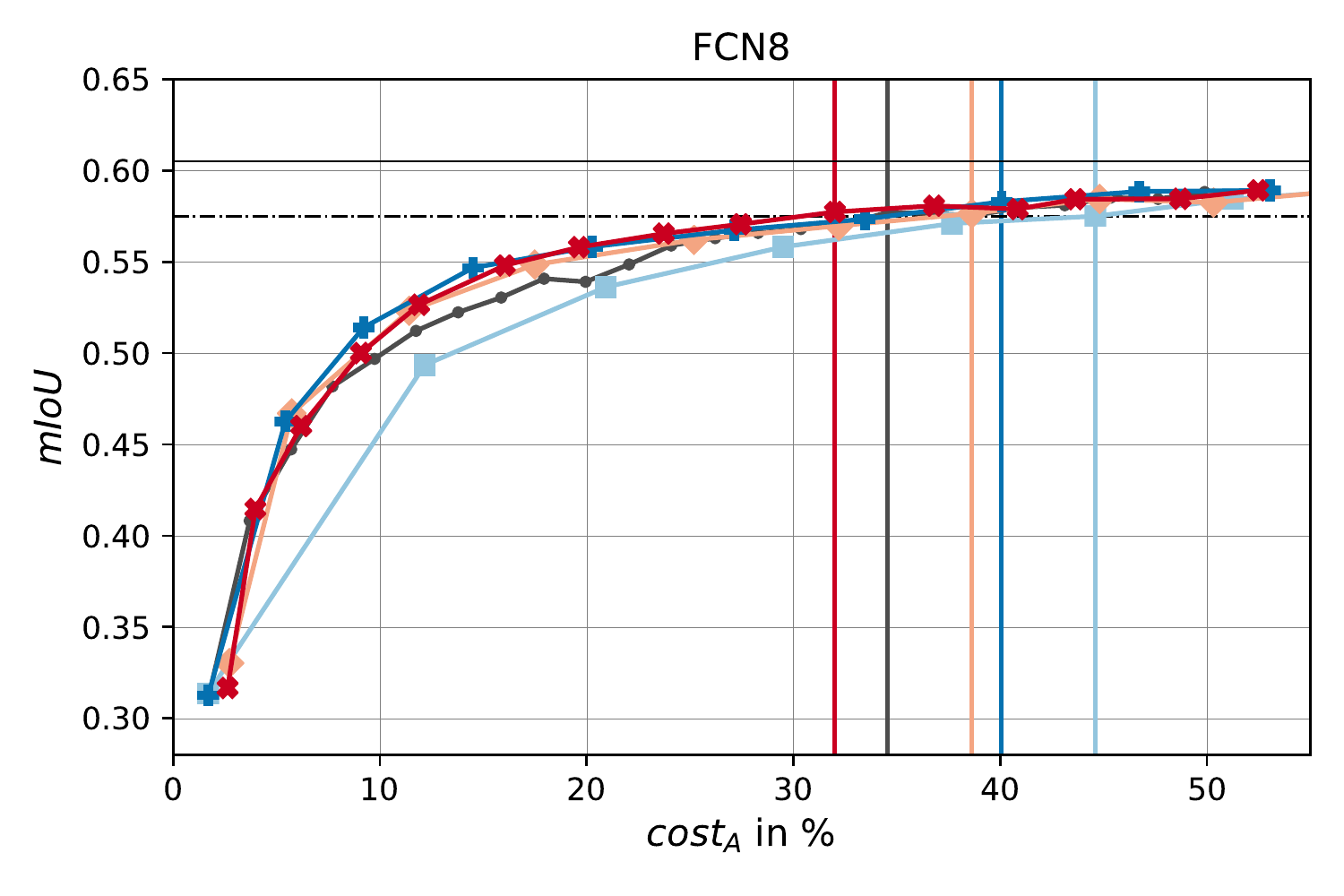} \hfill \includegraphics[width=.495\textwidth]{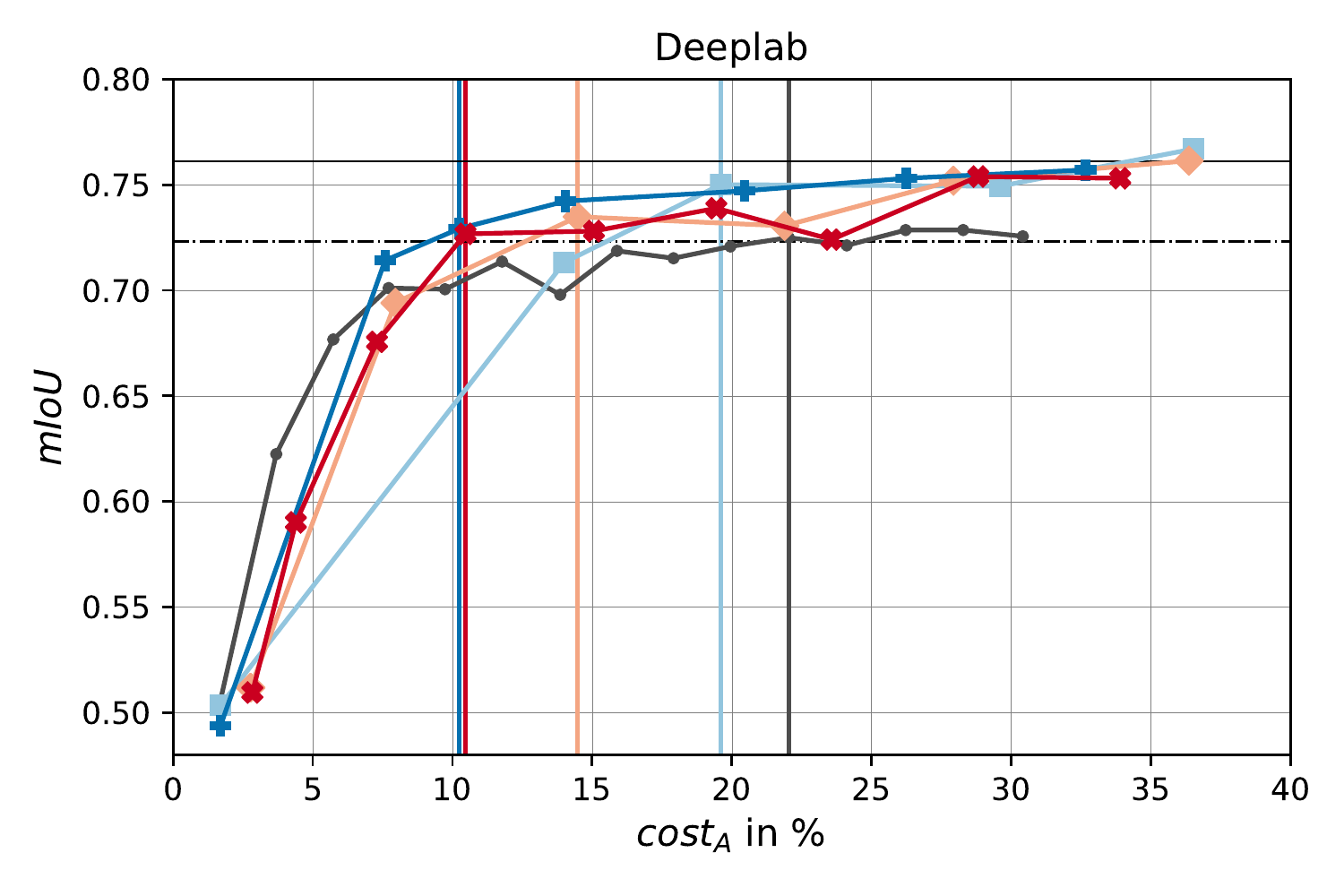}
}
\centerline{\includegraphics[width=.57\textwidth]{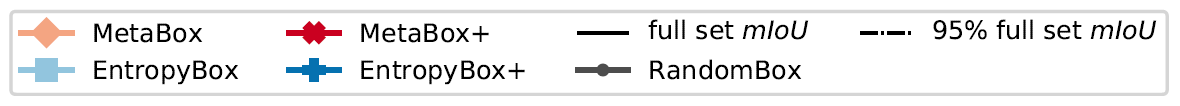}}
\caption{Summary of the results of the AL experiments with the FCN8 model  \emph{(top)} and the Deeplab model  \emph{(bottom)}. The costs are given in cost metric $\cA$. The vertical lines display where the $95\%$  full set $\mIoU$ are achieved.
% The dot lines (EntropyBox+*, MetaBox+*) the results if we use the true clicks which corresponds to a perfect cost estimation.
% MetaSeg based method's have more initial costs due to the data $\mathcal{M}$ required for training MetaSeg.
Each method's curve represents the mean over 3 runs.
}
\label{fig:experiments}
\end{figure*}

\paragraph{Robustness and Variance}
% \Cref{fig:experiments}
Considering \Cref{fig:experiments}, where the experiment\commentMR{s} for each CNN model are shown in one plot, we observe that all methods show a clear dependence on the CNN model. 
In our experiments with the FCN8 we also observe that results show only insignificant standard deviation over the different trainings. Hence we did not include a figure for this finding and rather focus on discussing the robustness \commentMR{of} the methods \outMR{applied}\commentMR{with respect} to the Deeplab model.

\begin{figure}
\centerline{
\includegraphics[width=.248\textwidth]{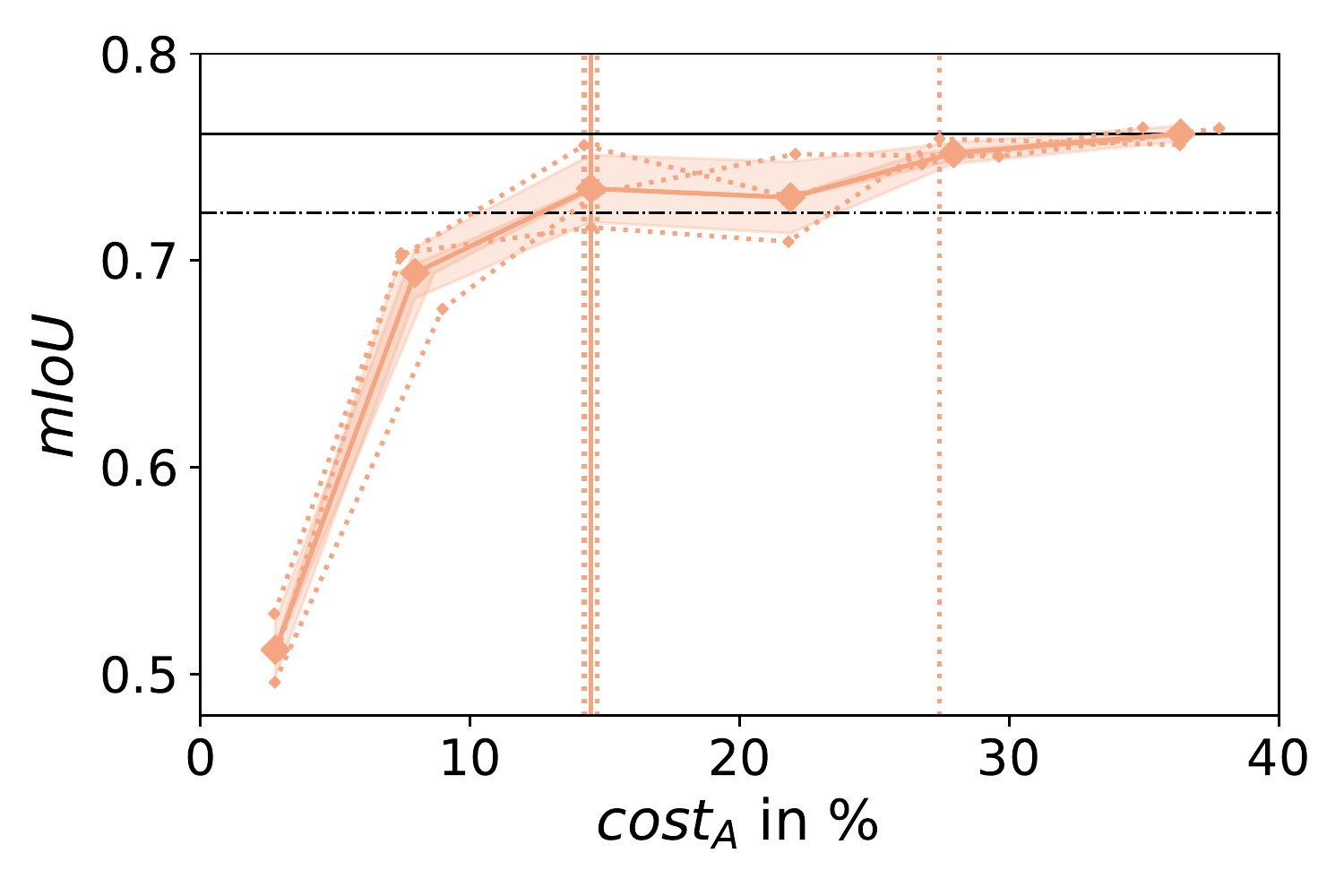} \hfill \includegraphics[width=.248\textwidth]{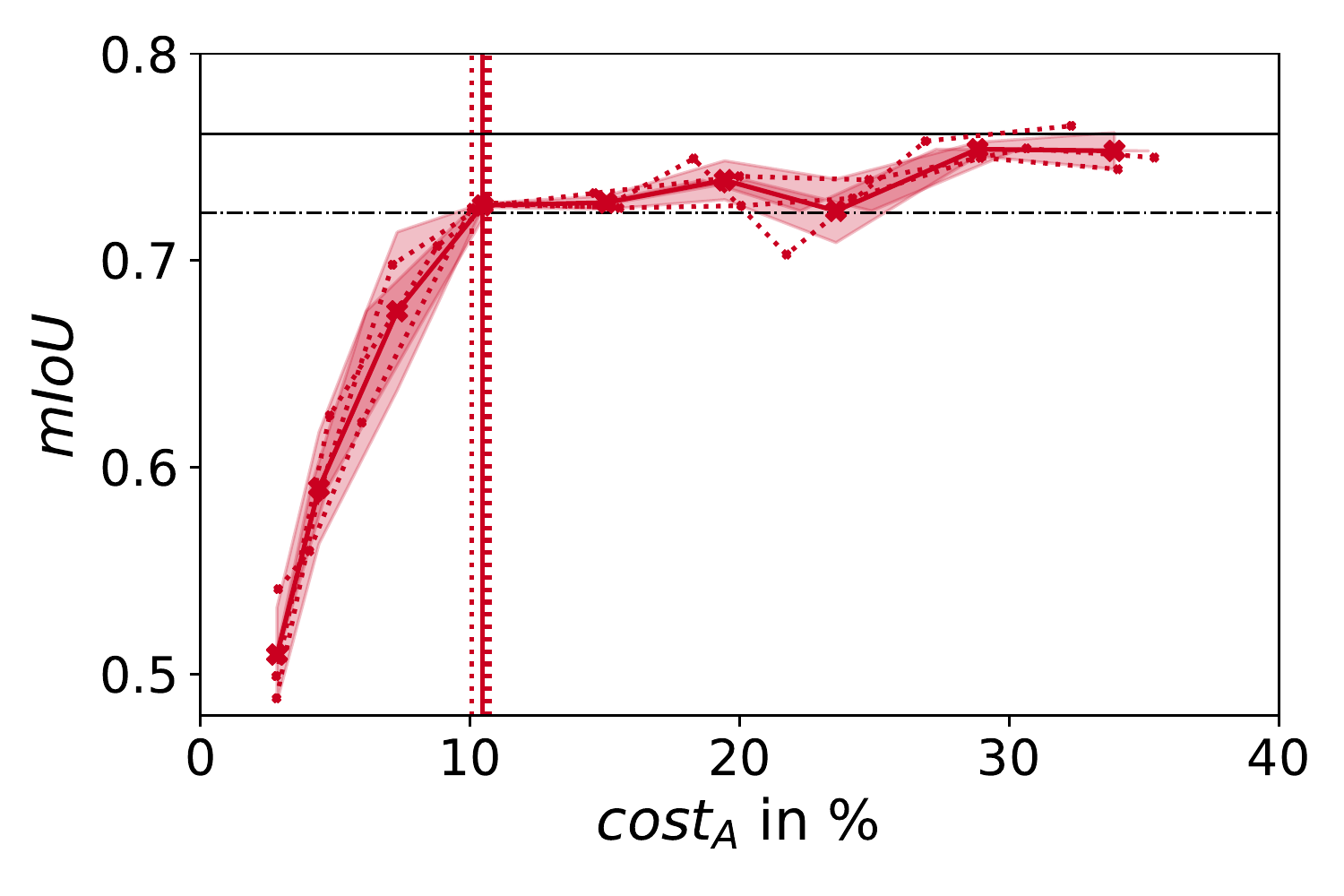}
}
\centerline{
\includegraphics[width=.248\textwidth]{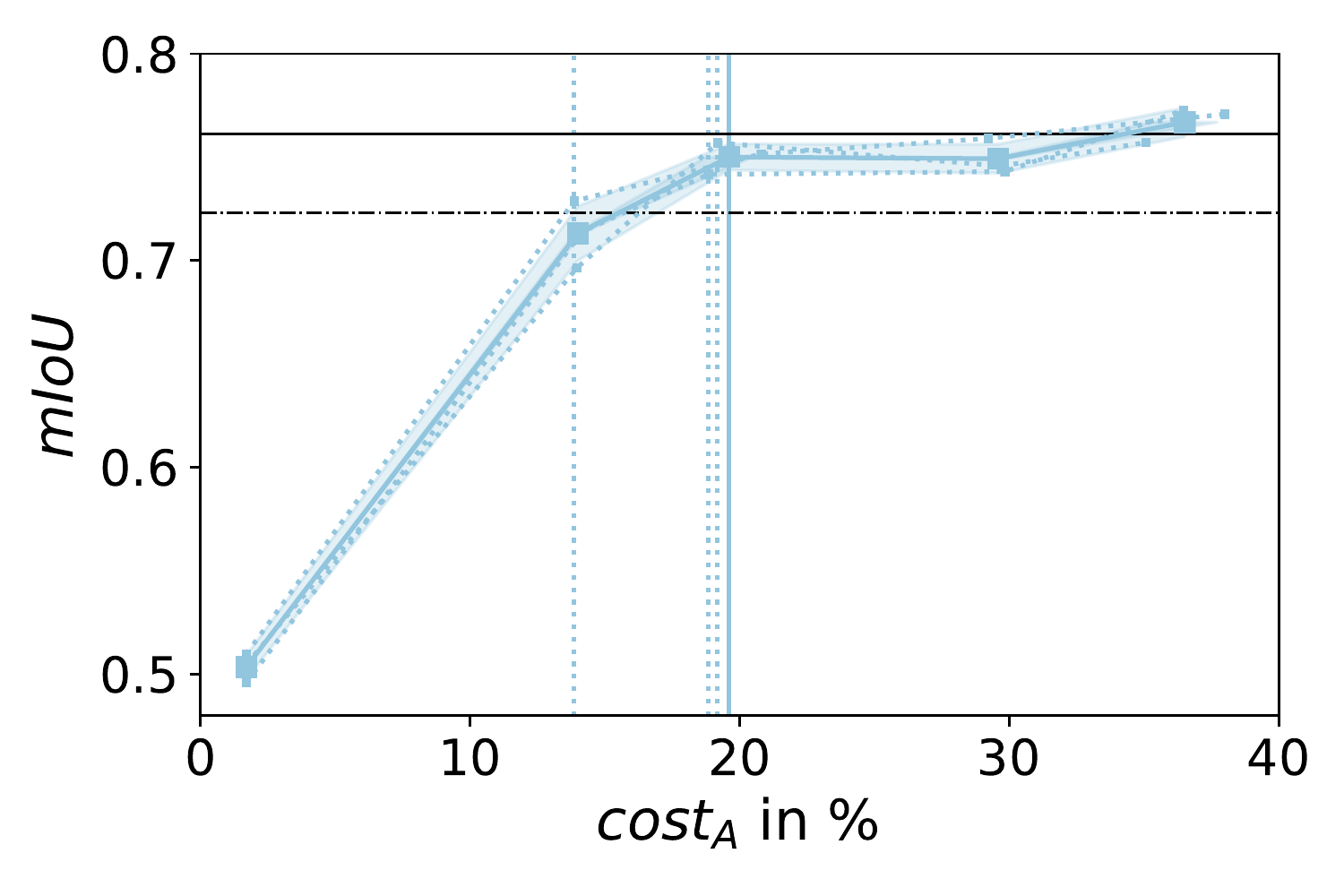} \hfill \includegraphics[width=.248\textwidth]{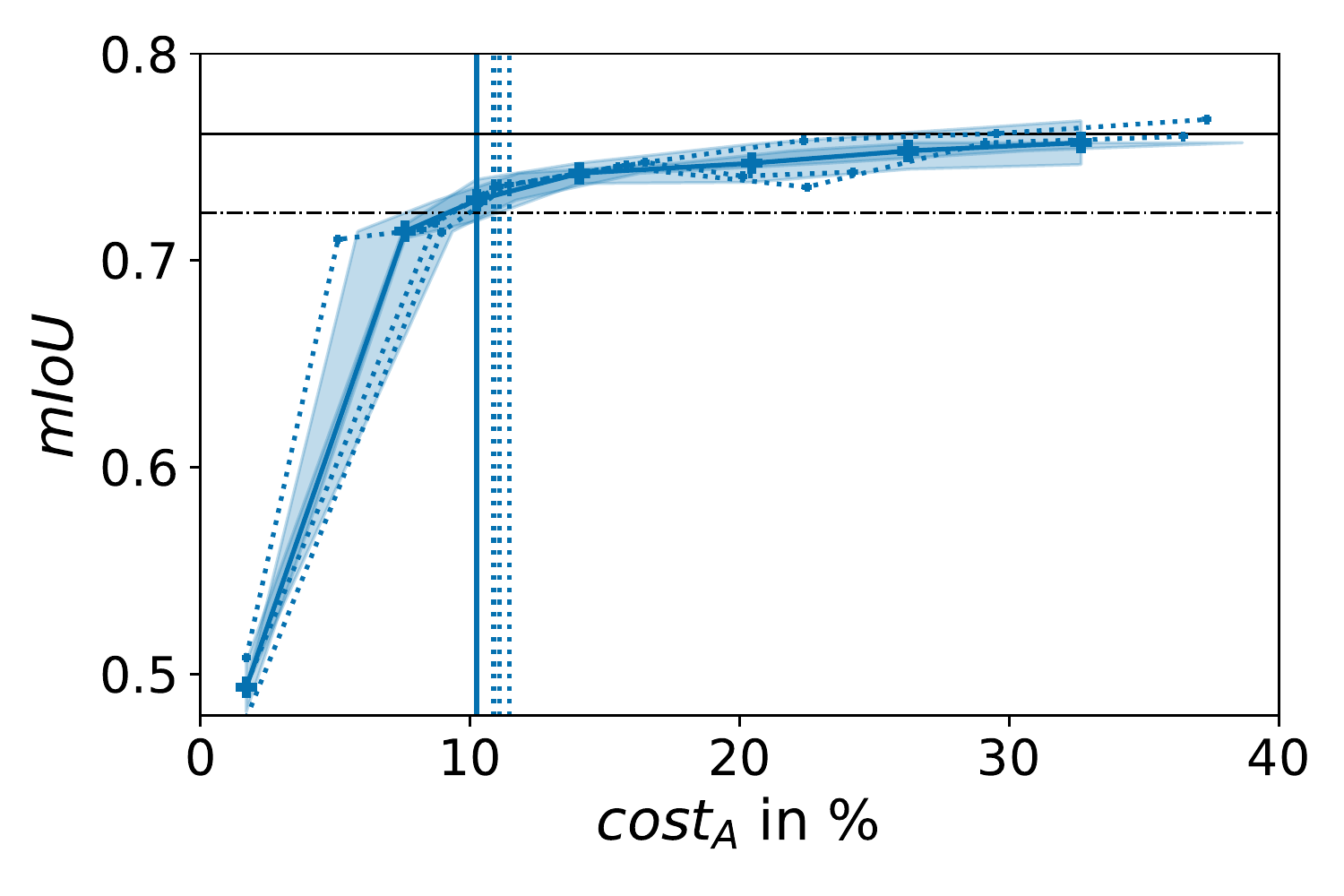} 
}
\centerline{
 \includegraphics[width=.248\textwidth]{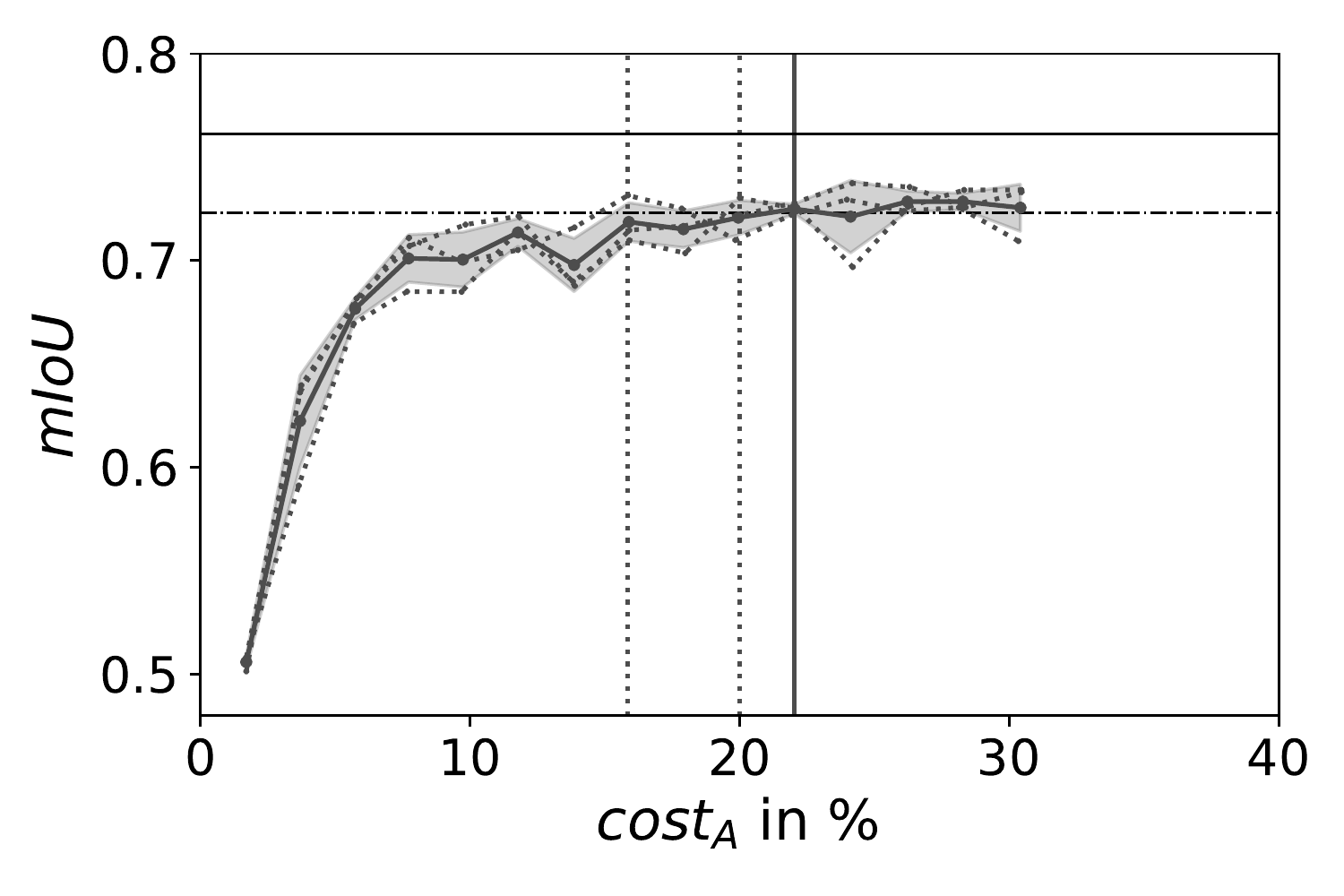} \hfill
\includegraphics[width=.248\textwidth]{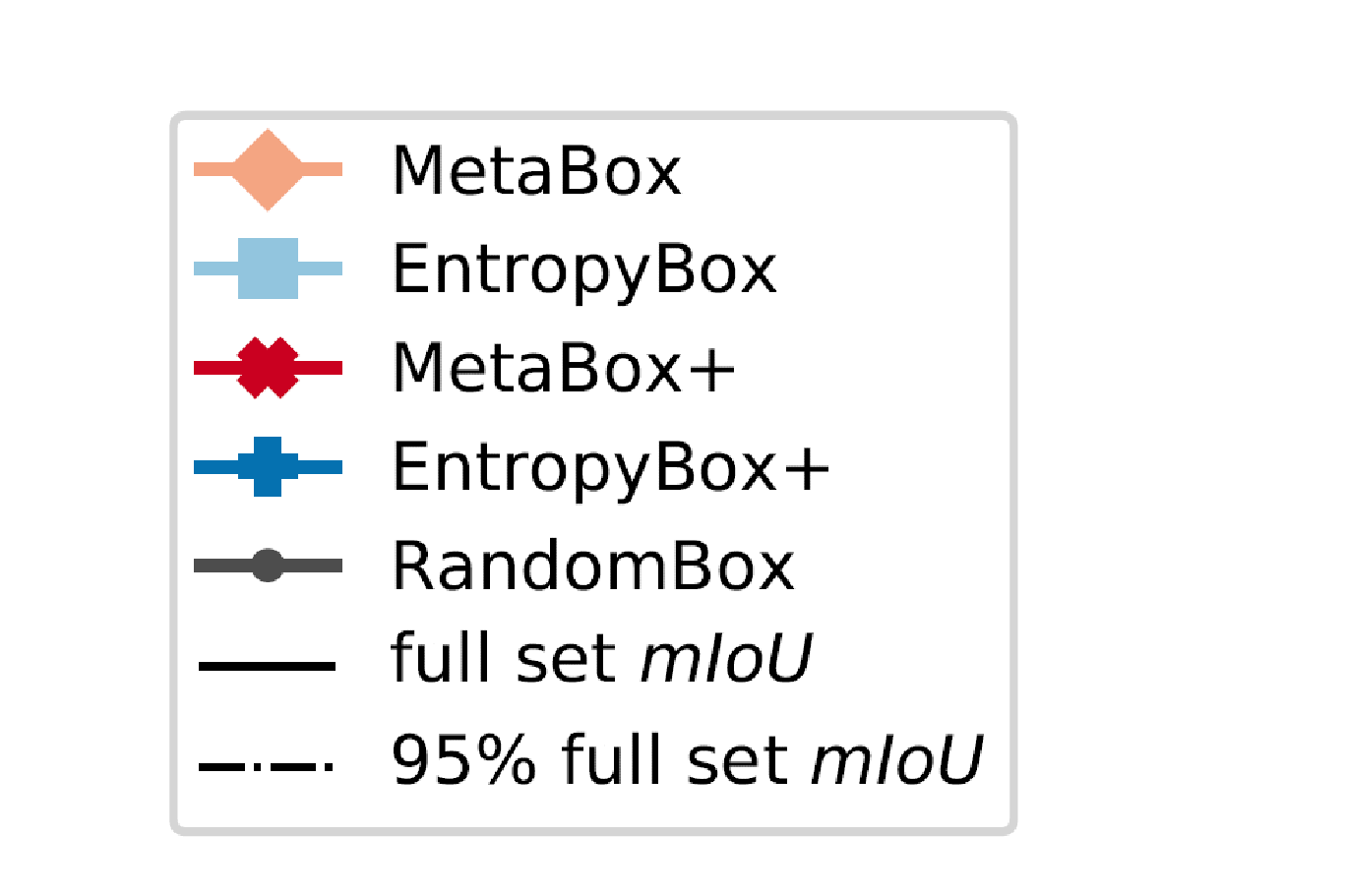}
}
\caption{Results of single AL experiments (consisting of 3 runs each) for each AL method with the Deeplab model. Costs are given in terms of cost metric $\cA$. In each plot, the single runs are given as dotted lines, the mean (over costs and $\mIoU$) as a solid line. The vertical lines show where each run achieves $95\%$  full set $\mIoU$.}
\label{fig:deeplab_experiments}
\end{figure}

For the Deeplab model, the methods show a significant standard deviation over trainings, especially the methods MetaBox, MetaBox+ and RandomBox, see \Cref{fig:deeplab_experiments}. In the first AL iterations, RandomBox rapidly gains performance at low costs. However, in the range of \outMR{the}$95 \%$ full set $\mIoU$ it rather fluctuates and only slightly gains performance. Beyond the $95 \%$ full set $\mIoU$ frontier, the methods MetaBox(+) and EntropyBox(+) still improve at a descent pace. MetaBox+ and EntropyBox+ nearly achieve the full set $\mIoU$ with approximately the same costs.

Furthermore, when investigating the variation of results with respect to two different CNN models, we observe that the discrepancy between the FCN8 and the Deeplab model is roughly $8$ \commentMR{percent points} smaller for MetaBox+ than for EntropyBox+. This shows that MetaBox+ \outPC{is} tends to be more robust with respect to the choice of CNN model.

% Plots for Deeplab

\paragraph{Comparison of cost metrics.}
In the evaluation above, we only consider the cost metric $\cA$. 
%Measuring the annotation effort with cost metric $\cB$, the costs are in general slightly lower.
A comparison of cost metrics for both CNN models is given in \Cref{tab:results}.
Note that EntropyBox+* and MetaBox+* refer to methods that are equipped with the true costs from the Cityscapes dataset. We elaborate further on this aspect in next paragraph.
Except for RandomBox, the required costs to achieve $95 \%$  full set $\mIoU$ is  
up to 3 \commentMR{percent points} lower when considering $\cB$ instead of $\cA$.
Considering the proportion of labeled pixels $\cP$ makes the costs seem significantly lower.
Noteworthily, for the FCN8 model EntropyBox requires only costs of $\cP = 10.08 \%$ while RandomBox does require costs of $\cP = 28.57 \%$, which is

% roughly a factor of 3 higher.
% Comparing this with $\cA=44.60\%$ it becomes clear, that these 10\% of the pixels in the dataset constitute to almost half of the actually required click work.
% This comparison \commentMR{highlights} the importance of cost measurement (definition of a cost metric) and that the annotation of image regions requires different human annotation effort. 

% Results and development of experiments as plots

% Results as table 
\begin{table*}

\begin{center}
\begin{tabular}{| c |c | c | c | c | c | c | c | c |}
\hline
CNN & Cost  & RandomBox & \multicolumn{3}{|c|}{EntropyBox} & \multicolumn{3}{|c|}{MetaBox} \\
\cline{3-9}
            model            & metric       &          &         & $+$     & $+^*$   &         & $+$     & $+^*$  \\ 
\hline \hline
\multirow{3}{*}{FCN8} & $\cA$  & $34.54$ & $44.60$ & $40.06$ & $\mathit{19.82}$ & $38.63$ & $\mathbf{32.01}$ & $\mathit{26.61}$  \\
                        & $\cB$  & $35.76$ & $40.74$ & $37.55$ & $\mathit{18.87}$ & $35.58$ & $\mathbf{30.85}$ & $\mathit{25.74}$  \\
                        & $\cP$  & $28.57$ & $\mathbf{10.08}$ & $13.45$ & $\mathit{10.08}$ & $12.77$ & $17.81$ & $\mathit{16.13}$  \\
\hline \hline
\multirow{3}{*}{Deeplab} & $\cA$  & $22.04$ & $19.61$ & $\mathbf{10.25}$ & $\mathit{10.95}$ & $14.48$ & $10.47$ & $\mathit{16.21}$  \\
                           & $\cB$  & $22.77$ & $17.92$ & $\mathbf{9.85}$ & $\mathit{10.60}$ & $13.56$ & $10.43$ & $\mathit{15.85}$  \\
                           & $\cP$  & $18.49$ & $5.04$ & $\mathbf{5.04}$ & $\mathit{6.72}$ & $6.05$ & $7.73$ & $\mathit{11.09}$  \\
\hline
\end{tabular}
{\caption{Annotation costs in $\%$ (row) produced by each method (column) to achieve $95 \%$  full set $\mIoU$. Cost metrics $\cA$ (\Cref{eq:cA}) and $\cB$ (\Cref{eq:cB}) are based on annotation clicks while cost metric $\cP$ (\Cref{eq:cP}) indicates the amount of labeled pixels. The costs \commentPC{with respect to the cost metric} of the best performing methods are highlighted. The strategies EntropyBox+* and MetaBox+* represent a hypothetical ``optimum'' by knowing the true costs. Each value was obtained as the mean over 3 runs.
} \label{tab:results}}
\end{center}
\end{table*}

\noindent
roughly a factor of 3 higher.
Comparing this with $\cA=44.60\%$ it becomes clear, that these 10\% of the pixels in the dataset constitute to almost half of the actually required click work.
This comparison \commentMR{highlights} the importance of cost measurement (definition of a cost metric) and that the annotation of image regions requires different human annotation effort. 

\paragraph{Click estimation.}
 
To evaluate our cost estimation (\Cref{cost_estimation}), we compare it to the provided clicks in the Cityscapes dataset by considering the latter as a ``perfect'' cost estimation.
That is, we supply EntropyBox and MetaBox with the true costs and term these methods EntropyBox+* and MetaBox+*. A comparison of the different click estimations and the true clicks is given in \Cref{tab:results}.
For the FCN8 model, the experiments show that knowing the true costs in most cases improves the results: EntropyBox+* produces costs of $\cA = 19.82 \% $. This is the half of the costs of EntropyBox+. MetaBox+* produces costs of $\cA = 26.61 $, which is $6$ \commentMR{percent points} less costs compared to MetaBox+. 
For the Deeplab model, using true rather than estimated costs do not lead to better results.
However, in terms of cost metric $\cA$, EntropyBox+* produces $1$ \commentMR{percent point} more costs then EntropyBox+. MetaBox+* produces even $6$ pp.\ more costs then MetaBox+. Similarly, we see such an increase also with respect to the other cost metrics $\cB$ and $\cP$\commentMR{.}

\section{Conclusion and Outlook} \label{sec:conclusion}

We have introduced a novel AL method MetaBox+, which is based on the estimated segmentation quality, combined with a practical cost estimation. 
We \outPC{also}compared MetaBox(+) to entropy based methods. \commentPC{Using a combination of entropy and our introduced cost estimation shows also \outMR{impressive}\commentMR{remarkable} results.}
\outPC{, which also show impressive results using our cost estimation.} Our experiments include in-depth studies for two different CNN models, comparisons of cost metrics, cost / click estimates, three different query types (Random, Entropy, MetaSeg) as well as a study on the robustness. The new methods MetaBox+ proposed by us lead to robust reductions in annotation cost, resulting in requiring $10$-$30\%$ annotation costs for achieving $95 \%$  full set $\mIoU$.
All our tests were conducted using \commentMR{a query function that minimizes} the product of two targets, i.e., minimizing the annotation effort and minimizing the estimated segmentation quality for a given query region. 
We leave the question open, whether a weighted sum of priorities instead of a product \Cref{eq:product} would lead to additional improvements of our methods.
\commentPC{Since each method produces different annotation costs per AL iteration, it \outMR{would}\commentMR{could} be of interest to vary the number of queried boxes (per AL iteration) or to start each experiment with some RandomBox iterations.}
Furthermore, it would be interesting to also incorporate pseudo labels, i.e., to label regions of high estimated quality with the predictions of the CNN model. Semi-supervised approaches remain a promising direction for further improvements and will be investigated in the future.

\bibliography{ecai}
\end{document}